\documentclass{article}

\usepackage{graphicx}

\usepackage[nonatbib,preprint]{neurips_data_2022}

\usepackage[utf8]{inputenc} 
\usepackage[T1]{fontenc}    
\usepackage{hyperref}       
\usepackage{url}            
\usepackage{booktabs}       
\usepackage{amsfonts}       
\usepackage{nicefrac}       
\usepackage{microtype}      

\usepackage{tikz}
\usepackage{comment}
\usepackage{amsmath,amssymb} 
\usepackage{color}
\usepackage{pifont}
\usepackage{mwe}
\usepackage{multirow}
\usepackage{multicol}
\usepackage{enumitem}

\usepackage{algorithm}
\usepackage{algpseudocode}
\usepackage{algorithmicx}
\usepackage{tabularx}
\usepackage{colortbl}

\PassOptionsToPackage{dvipsnames}{xcolor}
\usepackage[dvipsnames]{xcolor}

\title{PeRFception: Perception using Radiance Fields}

\author{
Yoonwoo Jeong$^{1\dagger}$\And Seungjoo Shin$^{1\dagger}$ \And Junha Lee$^{1\dagger}$ \And Christopher Choy$^2$ \And 
Animashree Anandkumar$^{2,3}$ \And Minsu Cho$^1$ \And Jaesik Park$^1$ \And
POSTECH$^1$~~~~~~NVIDIA$^2$~~~~~~Caltech$^3$
}

\begin{document}

\maketitle

\newcommand{\Eq}[1]  {Eq.\ (\ref{eq:#1})}
\newcommand{\Eqs}[1] {Eqs.\ (\ref{eq:#1})}
\newcommand{\Fig}[1] {Figure \ref{fig:#1}}
\newcommand{\Figs}[1]{Figures \ref{fig:#1}}
\newcommand{\Tbl}[1]  {Table \ref{tbl:#1}}
\newcommand{\Tbls}[1] {Tables \ref{tbl:#1}}
\newcommand{\Sec}[1] {Section \ref{sec:#1}}
\newcommand{\Secs}[1] {Sections \ref{sec:#1}}
\newcommand{\App}[1] {Appendix \ref{app:#1}}
\newcommand{\Etal}   {et al.}

\newcommand{\jaesik}[1]{\textcolor{cyan}{Jaesik: #1}\PackageWarning{Jaesik:}{#1!}}
\newcommand{\chris}[1]{\textcolor{blue}{Chris: #1}\PackageWarning{Chris:}{#1!}}
\newcommand{\yoonwoo}[1]{\textcolor{magenta}{Yoonwoo: #1}\PackageWarning{Yoonwoo:}{#1!}}
\newcommand{\seungjoo}[1]{\textcolor{violet}{Seungjoo: #1}\PackageWarning{Seungjoo:}{#1!}}
\newcommand{\junha}[1]{\textcolor{orange}{Junha: #1}\PackageWarning{Junha:}{#1!}}
\newcommand{\todo}[1]{\textcolor{red}{Todo: #1}\PackageWarning{TODO:}{#1!}}
\newcommand{\mcho}[1]{\textcolor{magenta}{Minsu: #1}}
\newcommand{\revision}[1]{\textcolor{black}{#1}}

\newcommand{\success}[1]{\textcolor{green}{Success: #1}}
\newcommand{\fail}[1]{\textcolor{green}{Fail: #1}}

\newcommand{\mathMesh}{\mathcal{M}}
\newcommand{\mathFace}{f}

\newcommand{\fix}{\marginpar{FIX}}
\newcommand{\new}{\marginpar{NEW}}
\newcommand{\xmark}{\ding{55}}%
\newcommand{\cmark}{\ding{51}}%

\newcolumntype{Y}{>{\centering\arraybackslash}X}

\newcommand{\our} {PeRFception-}
\newcommand{\ourdataset} {PeRFception dataset}

\definecolor{brickred}{rgb}{0.8, 0.25, 0.33}
\definecolor{bananayellow}{rgb}{1.0, 0.88, 0.21}
\definecolor{falured}{rgb}{0.5, 0.09, 0.09}

\begin{abstract}
    The recent progress in implicit 3D representation, i.e., Neural Radiance Fields (NeRFs), has made accurate and photorealistic 3D reconstruction possible in a differentiable manner.
    This new representation can effectively convey the information of hundreds of high-resolution images in one compact format and allows photorealistic synthesis of novel views. 
    In this work, using the variant of NeRF called Plenoxels, we create the first large-scale implicit representation datasets  for perception tasks, called the  \textbf{\ourdataset{}}, which consists of two parts that incorporate both object-centric and scene-centric scans for classification and segmentation. It shows a significant memory compression rate (96.4\%) from the original dataset, while containing both 2D and 3D information in a unified form. We construct the  classification and segmentation models that directly take as input this implicit format and also propose a novel augmentation technique to avoid overfitting on backgrounds of images. \revision{The code and data are publicly available in \url{https://postech-cvlab.github.io/PeRFception/}}. 
\end{abstract}

\section{Introduction}

Over the last few years, advances in implicit representations have demonstrated great accuracy, versatility, and robustness in representing 3D scenes by mapping low dimensional coordinates to the local properties of the scene, such as occupancy~\cite{mescheder2019occupancy,peng2020convolutional}, signed distance fields~\cite{park2019deepsdf,takikawa2021neural}, or radiance fields~\cite{mildenhall2020nerf,barron2021mipnerf,zhang2020nerf++}.
They offer several benefits that explicit representations (e.g., voxels, meshes, and point clouds) could not represent: smoother geometry, less memory space for storage, novel view synthesis with high visual fidelity, to name a few.
Thus, implicit representations have been used for 3D reconstruction~\cite{mescheder2019occupancy,peng2020convolutional,chen2018implicit_decoder,saito2019pifu}, novel view synthesis~\cite{mildenhall2020nerf,barron2021mipnerf,zhang2020nerf++,hedman2021baking,liu2020neural,muller2022instant,sitzmann2019scene,yu2021plenoxels,yu2021plenoctrees}, pose estimation~\cite{jeong2021self,Wang21arxiv_nerfmm,YenChen20arxiv_iNeRF, lin2021barf}, image generation~\cite{niemeyer2021giraffe,Schwarz20neurips_graf}, and many more. 

In particular, Neural Radiance Fields~\cite{mildenhall2020nerf} (NeRF) and many follow-up works~\cite{hedman2021baking,liu2020neural,muller2022instant,yu2021plenoxels,yu2021plenoctrees,sun2021direct} have shown that implicit networks can capture accurate geometry and render photorealistic images by representing a static scene as an implicit 5D function which outputs view-dependent radiance fields.
They use differentiable volumetric rendering, a scene geometry, and the view-dependent radiance that can be encoded into an implicit network using only image supervisions.
These components allow the networks to capture high fidelity photometric features, such as reflection and refraction in a differentiable manner unlike the conventional explicit 3D representations.

{\let\thefootnote\relax\footnotetext{${}\dagger$ Authors contributed equally to this work.}}

Given the success of the implicit representations, it is only natural to consider the implicit representation as one of the standard data representations for 3D and for perception.
However, these novel representations, which can capture a scene with high fidelity, have not yet been used for perception tasks such as classification and segmentation. 
One of the main reasons is that there is no large-scale dataset that allows training a perception system. Thus, in this work, we present the first large-scale implicit representation datasets to accelerate perception research.

Indeed, NeRFs have drawbacks that prevent the broad adoption of implicit representations as to the standard data format for 3D scenes and perception.
First, training an implicit network is slow and can take up to days. Inference (volumetric rendering) also can take minutes, limiting the use of NeRFs in real-time applications.
Second, the geometry and visual properties of a scene are implicitly encoded as weights in a neural network. These facts prevent an existing perception pipeline from processing the information directly.
Third, implicit features or weights are scene-specific and are not transferrable between scenes. However, for perception, channels or features must have a consistent structure, such as RGB channels for images. For instance, if the order of channels is different from an image to an image, the image classification pipeline would not work properly.

\begin{figure*}[!t]
\centering
\includegraphics[width=\textwidth]{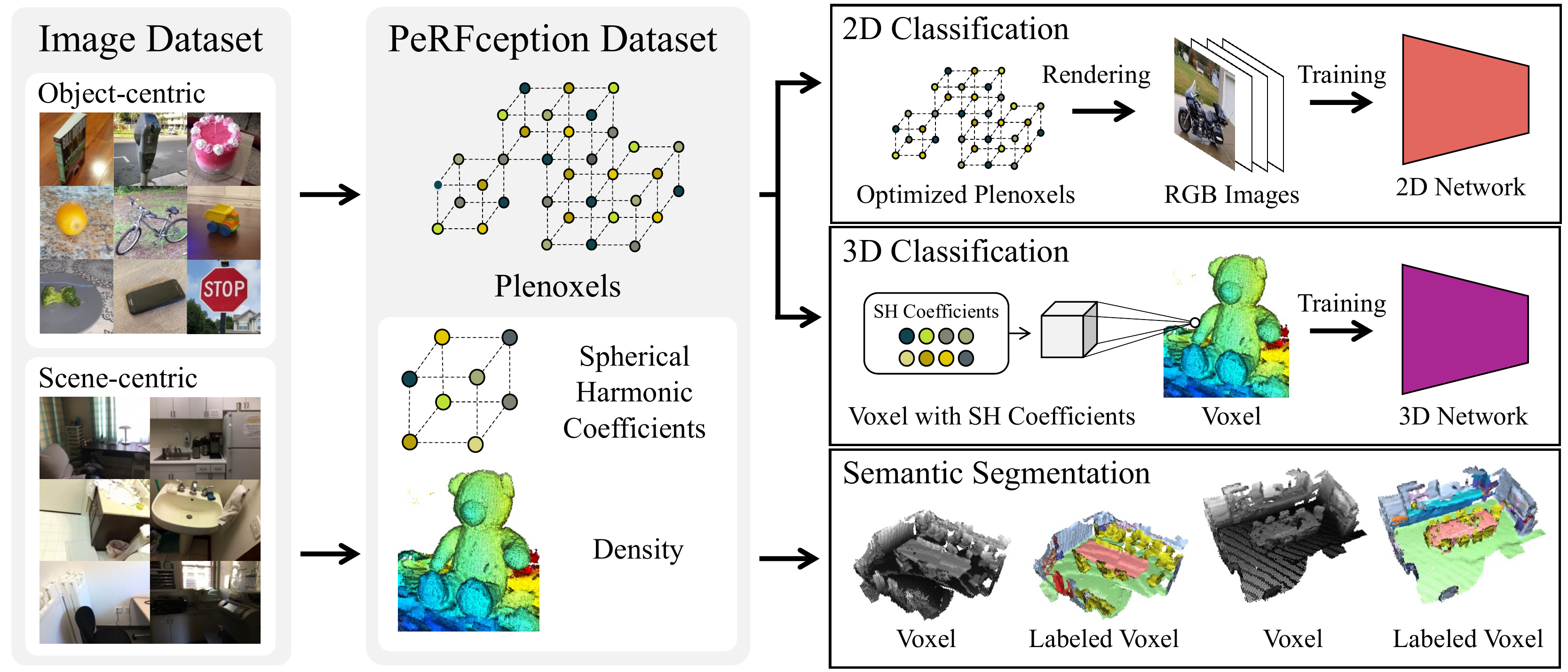}
\vspace{-4.0mm}
\caption{Overall illustration of \ourdataset{} with its applications. Our \ourdataset{} convey both visual (spherical harmonic coefficient) and geometric (density, sparse voxel grid) features in one compact format, it can be directly applied to various perception tasks, including 2D classification, 3D classification, and 3D segmentation.}
\label{fig:teaser}
\end{figure*}

Recent studies have resolved these limitations by adopting explicit sparse voxel grid geometry and basis functions for features.
First, to tackle the slow speed, many works propose to use the explicit sparse voxel geometry, which reduces the number of samples along a ray by skipping empty space~\cite{hedman2021baking,liu2020neural,muller2022instant,yu2021plenoxels,yu2021plenoctrees,sun2021direct}.
Second, instead of using the implicit representations of weights of a network, directly optimizing features~\cite{yu2021plenoxels,yu2021plenoctrees,sun2021direct} assigned to explicit geometry reduces the time to extract features from a network.
Lastly, for consistent features between scenes, which is crucial for perception or creating a scene with different objects in NeRF format, Yu et al.~\cite{yu2021plenoxels,yu2021plenoctrees} show that spherical harmonic coefficients can represent a scene as accurately as NeRFs while preserving consistent and structured features.
In particular, Plenoxels~\cite{yu2021plenoxels} satisfy all criteria for data representation which supports fast learning and rendering while maintaining a consistent feature representation for perception and composition of scenes.

In this work, we adopt Plenoxels as the primary format for perception tasks and create both object-centric and scene-centric environments.
We mainly use two image datasets and convert them into Plenoxels, the Common Object 3D dataset (CO3D)~\cite{reizenstein2021common} and ScanNet~\cite{dai2017scannet}, and name the converted datasets as \our CO3D and \our ScanNet, respectively.
As the size of Plenoxels can be extremely large, we present a few techniques to compress the size of data and hyperparameters for each setup to maximize the accuracy while minimizing the data size.

We use the \ourdataset{}s to train networks for 2D image classification, 3D object classification, and 3D semantic segmentation. 
We successfully train networks for each perception task, indicating that our datasets effectively convey 2D and 3D information together in a unified format. Moreover, we show that our representation allows more convenient background augmentation and sophisticated camera-level manipulation. 

We summarize our contributions as follows:
\begin{itemize}[noitemsep,topsep=0pt,parsep=2pt,partopsep=2pt,leftmargin=*]
\renewcommand{\labelitemi}{$\bullet$}
    \item We introduce the first large-scale implicit datasets that can be readily used in downstream perception tasks, including 2D image classification, 3D object classification, and 3D scene semantic segmentation.
    \item We conduct the first comprehensive study of visual perception tasks that directly process the implicit representation. The extensive experiments show that our datasets effectively convey the information for 2D and 3D perception tasks.
    \item We provide the ready-to-use pipeline to generate the implicit datasets with fully automatic processes. We expect this automatic process allows generating a very large scale 3D dataset in future. 
\end{itemize}

\section{Related Work}

\subsection{Neural Implicit Representations}

Representing a scene using an explicit representation such as voxels, meshes, or point clouds has been the most widely used format, but these are discrete and introduce discretization errors. Neural implicit representations, on the other hand, use a neural network to approximate the geometry or properties of a scene continuously~\cite{mescheder2019occupancy,peng2020convolutional,park2019deepsdf,Oechsle2019ICCV}.
Mildenhall~\Etal~\cite{mildenhall2020nerf} showed that neural radiance representation can generate high fidelity renderings with view-dependent illumination effects using multi-layer perceptrons.
Many of recent studies extend such implicit representation to dynamic scenes~\cite{park2021nerfies,Niemeyer2019ICCV,park2021hypernerf,pumarola2020d}, conditional generation~\cite{grf2020,devries2021unconstrained,hao2021GANcraft, gu2022stylenerf}, pose-free~\cite{jeong2021self,Wang21arxiv_nerfmm,lin2021barf}, and many more.
In particular, research on efficient rendering~\cite{hedman2021baking,muller2022instant,yu2021plenoctrees,sun2021direct} has been one of the major directions since volumetric rendering could take minutes. 
Hedman~\Etal~\cite{hedman2021baking} propose to create sparse voxel grids after training to accelerate rendering. Similarly, Plenoctree~\cite{yu2021plenoctrees} uses an octree data structure instead of sparse voxels for fast rendering. 
DVGO~\cite{sun2021direct} and Plenoxels~\cite{yu2021plenoxels} also adopt the sparse voxel structure, and improve both inference and training time. INGP~\cite{muller2022instant} proposes a multi-level hash encoding which enables the fast convergence. Recently, TensoRF~\cite{TensoRF} boosts both training and inference time by factorizing 3D radiance fields into lower dimensional vectors or matrices. 
In this work, we use Plenoxels for our data format since they have explicit geometry and consistent features in form of spherical harmonic coefficients. 

\subsection{3D Perception Datasets}

Over the last decade, many public large-scale datasets of real objects for 3D perception have been published thanks to the advances in commodity sensors. In this section, we cover such large-scale 3D datasets for objects and scenes.

ShapeNet~\cite{chang2015shapenet} and ModelNet~\cite{wu20153d} provide class and part annotations that are from synthetic CAD models. 
Early object-centric 3D datasets augment image datasets with 3D CAD model annotations. Pascal3D+~\cite{xiang2014beyond} and Objectron~\cite{xiang2016objectnet3d} contain 3D shapes that are matched with real-world 2D images containing objects; however, 3D models are chosen from approximately aligned 3D models, not precisely reconstructed from the corresponding 2D images.
Redwood~\cite{choi2016large} is a large-scale object-centric RGB-D scan video dataset, where only a few categories include 3D models and camera poses.
GSO~\cite{google2021gso} holds clear 3D models of real objects with textures, but missing physically rendered images.
\revision{3D-Future~\cite{fu20213d} provides synthetic CAD shapes with high-resolution informative textures developed by professional designers.}
CO3D~\cite{reizenstein2021common} provides large-scale object-centric videos with camera poses and high-quality point cloud models. They assess quality of reconstructed 3D shapes using human-in-the-loop validation and marked 5,625 point clouds as successfully reconstructed. 
Recently, ABO~\cite{collins2021abo} offers a dataset consisting of household object images and high-quality 3D models with 4K texture maps and full-view coverage. Professional artists manually designed its high-quality spatially-varying Bidirectional Reflectance Distribution Functions (BRDFs), indicating that the data generation processes were not fully automatic. We summarize the details of the aforementioned datasets in Table~\ref{tab:main_data_spec_benchmark}.

\begin{table}[!htb]
    \centering
    \caption{Specs for publicly available 3D datasets. ``Real" denotes whether the objects are from real-world images
    , ``Full 3D" for the availability of 3D geometries for all the objects, and ``Multi-view" for multi-view images and corresponding real-world catalog images. \textcolor{bananayellow}{$\blacktriangle$} is marked when the corresponding information is partially provided. }
    \resizebox{\textwidth}{!}{
    
    \begin{tabularx}{1.2\textwidth}{ c *{6}{Y} }
        \toprule
        Dataset & \# Classes & \# Objects & Real & Full 3D & Multi-view \\
        \midrule
        ShapeNet\cite{chang2015shapenet} & 55 & 51K & \textcolor{falured}{\xmark} & \textcolor{teal}{\checkmark} & \textcolor{teal}{\checkmark} \\
        ModelNet\cite{wu20153d} & 40 & 128K & \textcolor{falured}{\xmark} & \textcolor{teal}{\checkmark} & \textcolor{falured}{\xmark} \\
        Pascal3D+\cite{xiang2014beyond} & 12 & 36K & \textcolor{teal}{\checkmark} & \textcolor{falured}{\xmark} & \textcolor{falured}{\xmark} \\
        Redwood\cite{choi2016large} & 9 & 2K & \textcolor{teal}{\checkmark} & \textcolor{falured}{\xmark} & \textcolor{teal}{\checkmark} \\
        Objectron\cite{xiang2016objectnet3d} & 9 & 15K & \textcolor{teal}{\checkmark} & \textcolor{bananayellow}{$\blacktriangle$} & \textcolor{bananayellow}{$\blacktriangle$} \\
        GSO\cite{google2021gso} & \textcolor{falured}{\xmark} & 2K & \textcolor{teal}{\checkmark} & \textcolor{teal}{\checkmark} & \textcolor{falured}{\xmark} \\
        3D-Future\cite{fu20213d} & 8 & 2K & \textcolor{falured}{\xmark} & \textcolor{teal}{\checkmark} & \textcolor{falured}{\xmark} \\
        ABO\cite{collins2021abo} & 98 & 8K & \textcolor{bananayellow}{$\blacktriangle$} & \textcolor{teal}{\checkmark} & \textcolor{teal}{\checkmark} \\
        CO3D\cite{reizenstein2021common} & 51 & 19K & \textcolor{teal}{\checkmark} & \textcolor{bananayellow}{$\blacktriangle$} & \textcolor{teal}{\checkmark} \\
        \rowcolor{yellow!40}
        \midrule
        \our-CO3D & 51 & 19K & \textcolor{teal}{\checkmark} & \textcolor{teal}{\checkmark} & \textcolor{teal}{\checkmark} \\
        \bottomrule
    \end{tabularx}
    }
    \label{tab:main_data_spec_benchmark}
\end{table}

\noindent Many scene-centric 3D datasets use depth sensors to scan a section or an entire room and create dense annotations.
SUN RGB-D~\cite{song2015sun} collected 13,355 RGB-D images, which are densely annotated with 2D polygons and 3D bounding boxes. However, it does not include camera parameters, which is essential information for surface reconstruction.
NYUv2~\cite{silberman2012indoor} initially sparked interest for 3D scene understanding, with 464 indoor scans, 1,449 frames of which are annotated with 2D polygons for semantic segmentation.
SUN3D\cite{xiao2013sun3d} is comprised of 415 RGB-D indoor video sequences in 254 different spaces; only eight sequences are annotated.
Each sequence was captured densely, with a large number of frames collected.
2D-3D-S~\cite{armeni20163d,armeni2017joint} is an instance-level annotated large-scale indoor scene dataset.
It offers diverse modalities of six indoor scenes in RGB images, depth maps, surface normals, 3D meshes, and point clouds.
Currently, ScanNet~\cite{dai2017scannet} is the most popular large-scale indoor scene dataset that collected instance-level annotated 1,513 scans of RGB-D images and 3D data.
Matterport3D~\cite{chang2017matterport3d} contains large-scale RGB-D images annotated with surface and semantic information.
In particular, it covers a wide area of 90 building-scale scenes by capturing panoramic views, but it does not provide annotations.
Replica~\cite{straub2019replica} is a small but high-quality surface-annotated indoor scene reconstruction database.
In this paper, we use two datasets, CO3D and ScanNet, to cover both object- and scene-centric dataset respectively.

We select CO3D for object-centric dataset since it consists of camera-annotated real-world images and has sufficient number of classes. 
In addition, we use ScanNet since it is one of the most popular 3D indoor dataset providing adequate number of data with rich annotations.

\subsection{3D Perception Models}
Unlike perceptions in the 2D domain, where the image is the de facto standard representation, there is no canonical representation for spatial 3D data. 
Existing explicit representations, such as voxels, meshes, and point clouds, target different aspects of data and have pros and cons. We categorize methods into two groups based on the input representation. 
Point-based methods~\cite{qi2017pointnet,qi2017pointnet++} directly consume the continuous 3D coordinates of point clouds or meshes using MLP and continuous/graph convolutions. 
Recent studies have tried to define custom convolution layers~\cite{mao2019interpolated,tatarchenko2018tangent,thomas2019kpconv,wu2019pointconv,xu2021paconv} upon the continuous coordinate space, or non-local operations~\cite{wang2019dynamic,yan2020pointasnl,zhao2021point}. 
Overall, these methods exhibit fast and simple processing, but it often requires a large computational cost due to neighbor search.

On the other hand, voxel-based methods discretize input coordinates into voxels, which introduces small quantization errors but allows fast neighbor lookup using a data structure. Specifically, recent advances in spatial sparse convolutions~\cite{graham2014spatially,tang2020searching,choy20194d} that operates on sparse voxels utilize an efficient GPU hash table and require small memory footprint for neighbor search.
It has shown successful adoption in many perception tasks, including semantic segmentation~\cite{tang2020searching,choy20194d,park2022fast}
, object detection~\cite{gwak2020generative}, representation learning~\cite{choy2019fully}, and registration~\cite{choy2020deep,lee2021deephough,predator,choy2020high}.
We use the spatial sparse convnets to create the first perception network on our \ourdataset{}s due to its scalability 
in terms of memory footprint and computational cost.
\section{Preliminary}

Yu~\Etal~\cite{yu2021plenoxels} proposed a novel scene representation called Plenoxels that combines a sparse voxel grid for coarse geometry with spherical harmonic coefficients for radiance fields. Unlike conventional MLP-based NeRFs ~\cite{mildenhall2020nerf,barron2021mipnerf,hedman2021baking,liu2020neural}, which use a single neural network to represent an entire scene, Plenoxels optimize coefficients of spherical harmonic in each non-empty voxel independently, which uses the same differentiable model for volumetric rendering described in NeRF~\cite{mildenhall2020nerf} as follows:
\begin{equation}
\hat{C}(\mathbf{r}) = \sum_{ i=1}^N T_i (1 - \exp(-\sigma_i \delta_i))\mathbf{c}_i \text{,~~~where } T_i = \exp(-\sum_{j=1}^{i-1} \sigma_j \delta_j),
\end{equation}
where $T_i$ denotes light transmittance of a $i$-th sample, $\sigma$ is opacity, $\mathbf{c}$ is color, and $\delta$ is distance to the next sample on a ray $\mathbf{r}$. Plenoxels lookup the stored densities and spherical harmonic coefficients in the sparse voxel grids. For scenes with background, Plenoxels also use the lumisphere background representation to render the backgrounds.

We modified the official Plenoxels implementation to set the initial grid properly and hyperparameters. As the size of Plenoxels can be extremely large, we present a few techniques to compress the size of data and hyperparameters. More implementation details are in Section.~\ref{subsec:co3d_generation}, Section.~\ref{subsec:scannet_generation}, and the appendix.

\section{Generating \ourdataset{}}

We generate two datasets \our{CO3D} and \our{ScanNet} to train perception networks. 

\begin{table}[!t]
    \centering
    \caption{Specs of CO3D and ScanNet, and our \our CO3D and \our ScanNet. \textbf{SH} denotes spherical harmonic coefficients, \textbf{D} for densities, and \textbf{C} for diffused color, "pcd" for point cloud. 3D-BG marks whether the 3D representation includes backgrounds of scenes. We note the number of frames in our proposed datasets as $\infty$ since our data representation is feasible to render frames from infinitely many camrea intrinsics and extrinsics.}
    \resizebox{\textwidth}{!}{
    \begin{tabular}{cccccccc}
        \toprule
        Dataset & \# Scenes & \# Frames & 3D Shape & Features & 3D-BG & Memory & Memory(Rel)\\
        \midrule
        CO3D & 18.6K & 1.5M & pcd & \textbf{C} & \textcolor{falured}\xmark & 1.44TB & $\pm0.00\%$ \\
        \rowcolor{yellow!40}
        \our CO3D & 18.6K & $\infty$ & voxel & \textbf{SH},\textbf{D} & \textcolor{teal}\cmark & 1.33TB & $-6.94\%$\\
        \midrule
        ScanNet & 1.5K & 2.5M & pcd & \textbf{C} & \textcolor{falured}\xmark & 966GB & $\pm0.00\%$ \\
        \rowcolor{yellow!40}
        \our ScanNet & 1.5K & $\infty$ & voxel & \textbf{SH},\textbf{D} & \textcolor{teal}\cmark & 35GB & $-96.4\%$ \\
        \bottomrule
    \end{tabular}
    }
    \label{tab:main_data_spec_co3d}
\end{table}

\begin{figure*}[!t]
\centering
\vspace{-4.0mm}
\includegraphics[width=\textwidth]{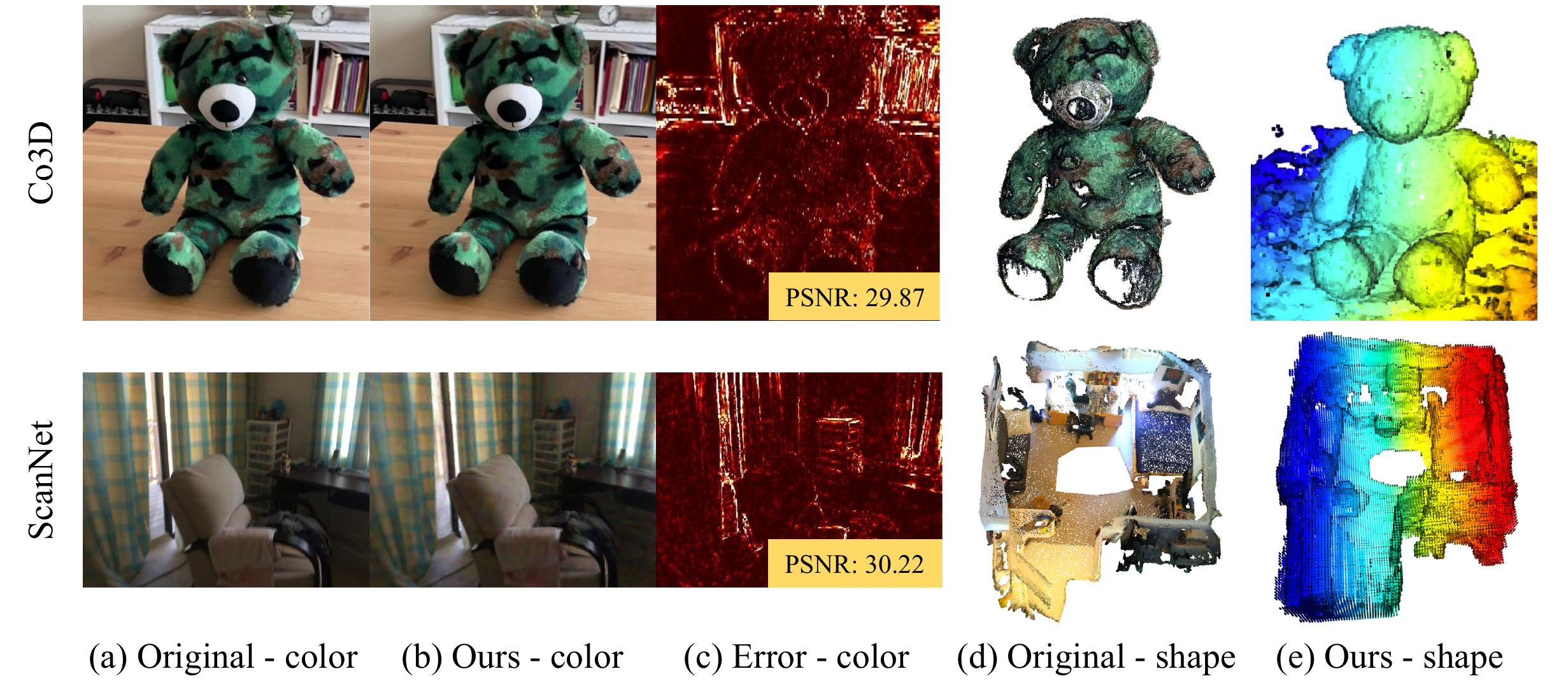}
\vspace{-5.0mm}
\caption{Visualization of a few example data of original datasets and our PeRFception datasets. From the source images and corresponding parameters, we successfully construct PeRFception datasets with both accurate geometry and photorealistic rendering. \revision{The color used in (e) is for visualization.}}
\label{fig:main_data_visualize}
\end{figure*}

\subsection{\our CO3D}
CO3D~\cite{reizenstein2021common} is a large-scale object-centric dataset that contains multi-view observations of objects.
It contains 18,669 annotated videos with a total 1.5 million of camera-annotated frames and 50 classes from MS-COCO~\cite{lin2014microsoft}, and images crowd-sourced from Amazon Mechanical Turk (AMT). 
It also provides reconstructed pointclouds, generated by pretrained instance segmentation algorithm~\cite{kirillov2019pointrend} and COLMAP~\cite{schonberger2016structure}. 
Although reconstructing depth and pointclouds is automatic, its generation step still requires human-in-the-loop validation. 
When the amount of data increases, this human-in-the-loop validation becomes unsuitable. On the other hand, our 3D dataset generation does not require manual verification since image reconstruction qualities on unseen views are used as a proxy for reconstruction quality.
We compare specs of the original CO3D and \our{CO3D} in Table~\ref{tab:main_data_spec_co3d}. 

\textbf{Data Generation.}
\label{subsec:co3d_generation}
We use the official implementation of Plenoxels~\cite{yu2021plenoxels} with a slight modification to the default configuration.
We reduce the resolution of the background lumisphere from 1,024 to 512 and the number of background layers from 64 to 16.
For sharper surface, we set the lambda sparsity value to $10^{-10}$, 10 times larger than the default configuration. 
A voxel grid is initialized with $128^3$ resolution and trained for 25,600 iterations. Then, it is upsampled once to $256^3$ resolution and trained for further 51,200 iterations.
Before saving the data, we quantize the trained parameters to unsigned 8-bit integers to minimize for storage except for density values. 
For each scene, we first filter out defective images and uniformly sample 10\% of the images as the test set to assess the rendering quality.
The quantitative and qualitative results of the rendering quality are reported in Table~\ref{tab:render_qual} and Figure~\ref{fig:main_data_visualize}.
More details are in the appendix.

\subsection{\our ScanNet}
ScanNet is a 3D-scanned indoor dataset that captures more than 1.5K indoor scenes with the commercial RGB-D sensors. It provides 3D reconstructed point clouds of scenes with semantic labels containing 20 common object classes, as well as the raw RGB-D frames with corresponding semantic masks and camera parameters. In our experiment, we follow the official data split and report the numbers on the validation split since the test set annotations are not publicly available. 

\textbf{Data Generation.} 
\label{subsec:scannet_generation}
ScanNet videos are captured using handheld cameras where auto-exposure option is held. So, a fair number of frames contain motion blur which could lead to poor scene geometries. 
\revision{In practice, we generate the batches of rays before training and load them in CPU memory for efficient memory bandwidth utilization during training. Since the number of frames for each scene in ScanNet varies, we use uniformly-sampled 1,500 image frames at most.} 
For the scenes with fewer than 1,500 images, we use them all after filtering out blurry images with a low variance of Laplacian~\cite{pertuz2013analysis}. 
Another characteristic of ScanNet is that, unlike object-centric datasets where cameras face inward, the images are captured from inside a room facing outward. 
These result in fewer images observing the same part of the space, which results in poor reconstruction of Plenoxel's geometry on ScanNet dataset. Specifically, the Plenoxel reconstruction artificially creates an excessive number of voxels in the empty space (i.e. floaters) to minimize the image reconstruction loss.

Instead, to supply an additional geometric prior to Plenoxel training, we initialize the voxel grid using the unprojected depth maps provided in ScanNet rather than starting from the dense voxel grid.
However, since the provided depth maps of ScanNet are contaminated with noisy observations, we incorporate the connected component analysis to filter out the disconnected outlier points in the unprojected point clouds.
This leads to stable and more accurate reconstruction and does not excessively generate floaters to minimize the rendering loss.
The resulting \our ScanNet dataset occupies only 35GB in disk whereas the original video streams of ScanNet requires about 966GB disk space.
This is a significant compression rate (96.4\%), which emphasizes the accessibility of our representation as a dataset.
Detailed dataset specs of the original ScanNet and \our{ScanNet} are reported in Table~\ref{tab:main_data_spec_co3d}. We report the rendering quality on Table~\ref{tab:render_qual} and visualize the qualitative novel view renderings on Figure~\ref{fig:main_data_visualize}. 
\begin{table}[!tb]
    \caption{Overall rendering qualities of \our CO3D and \our ScanNet on test set. For class-wise rendering scores are reported in the appendix. 
    }
    \centering
    \resizebox{\textwidth}{!}{
    \begin{tabular}{cccccccc}
        \toprule
        Dataset & PSNR($\uparrow$) & SSIM($\uparrow$) & LPIPS\footnotemark{}($\downarrow$) & \revision{Train Time (m)} & PSNR $>$ 15& PSNR $>$ 20 & PSNR $>$ 25 \\
        \midrule
        \our CO3D & 28.82 & 0.8564 & 0.3451 & \revision{21.6} & 99.8$\%$ & 98.2$\%$ & 87.3$\%$ \\
        \midrule
        \our ScanNet & 23.42 & 0.7470 & 0.4637 & \revision{11.3} & 99.7\% & 73.9\% & 40.4 \% \\
        \bottomrule
    \end{tabular}
    \vspace{2mm}
    }
    \label{tab:render_qual}
\end{table}


\footnotetext{The LPIPS metric sometimes generates ``nan" although their visual qualities are great enough. Only 0.01\% of scenes are noted as failure.}

\section{Experiment}

We benchmark popular 2D image classification, 3D object classification, and 3D segmentation networks to demonstrate that our unified data format can be used for various perception tasks. 

\subsection{Classification on \our{CO3D} dataset}
\label{subsec:3d_cls}
CO3D provides multi-view images of objects and 51 class labels for classification. We use the same class labels for classification of \our{CO3D} dataset.
We adopt a few classification models for our dataset for both 2D~\cite{he2016deep} and 3D classification~\cite{choy20194d}.
We split the dataset into the train, validation, and test set by scenes since the original CO3D does not provide such splits.
We use $10\%$ of the scenes for validation set and $10\%$ for test set in each class. We use the same splits for 2D and 3D classification.

\subsubsection{Implementation Details}

All the 2D classification models are trained with the cross-entropy loss with the weight decay factor $10^{-4}$. Following the recommendations from ~\cite{he2019bag}, we utilize the label smoothing with $\epsilon = 0.005$, remove bias decay, and initialize weights of the batch normalization layers on the residual connections to 0. We use the SGD optimizer with momentum 0.9 and trained for 500 epochs with a batch size 64. For 50 epochs, we linearly warmed up the learning rate from 0 to 0.1 and decayed it using the cosine annealing scheduler. It takes up to a day for training using a single RTX 3090 GPU. We have benchmarked variants of ResNet (ResNet18, ResNet34, ResNet50, ResNet101, ResNet152), ResNext~\cite{xie2017aggregated} (ResNext50, ResNext101), and WideResNet~\cite{zagoruyko2016wide} (WideResNet50, WideResNet101) networks.

For 3D classification, we train 3D version of ResNets~\cite{he2016deep} with varying depths that are implemented with spatially sparse convolutional layers~\cite{tang2020searching,choy20194d}. 
These networks directly take sparse voxels from Plenoxels as input. 
The Plenoxels consist of two components: coordinates of sparse voxels and their features (spherical harmonic coefficients and density values).
To demonstrate the efficacy of such features in perception task, we train the networks by providing different input features. 
We use the SGD optimizer and set the initial learning rate as 0.1 for all experiments and decay it with the cosine annealing scheduler for 100K iterations with batch size 16 on a single RTX 3090 GPU.
We augment the input data with both geometric augmentation (random rotation, coordinate dropout, random horizontal flip, coordinate uniform translation, and random scaling) and feature-level augmentation, random feature jittering. Further implementation details are in the appendix. 

\textbf{Background Augmentation.}
Plenoxels use both sparse voxels and lumispheres to render foregrounds and backgrounds respectively. In other words, we can render each of them separately or manipulate them to create various augmentations.  Specifically, we create a novel augmentation that substitutes the background in a scene with backgrounds from other scenes while preserving the foreground object. We describe the composition of foregrounds and randomly selected backgrounds in the appendix. In addition, we visualize several background augmentation examples in the appendix. 

\subsubsection{2D Classification on \our{CO3D}}

\begin{figure}[!t]
\centering
\includegraphics[width=\textwidth]{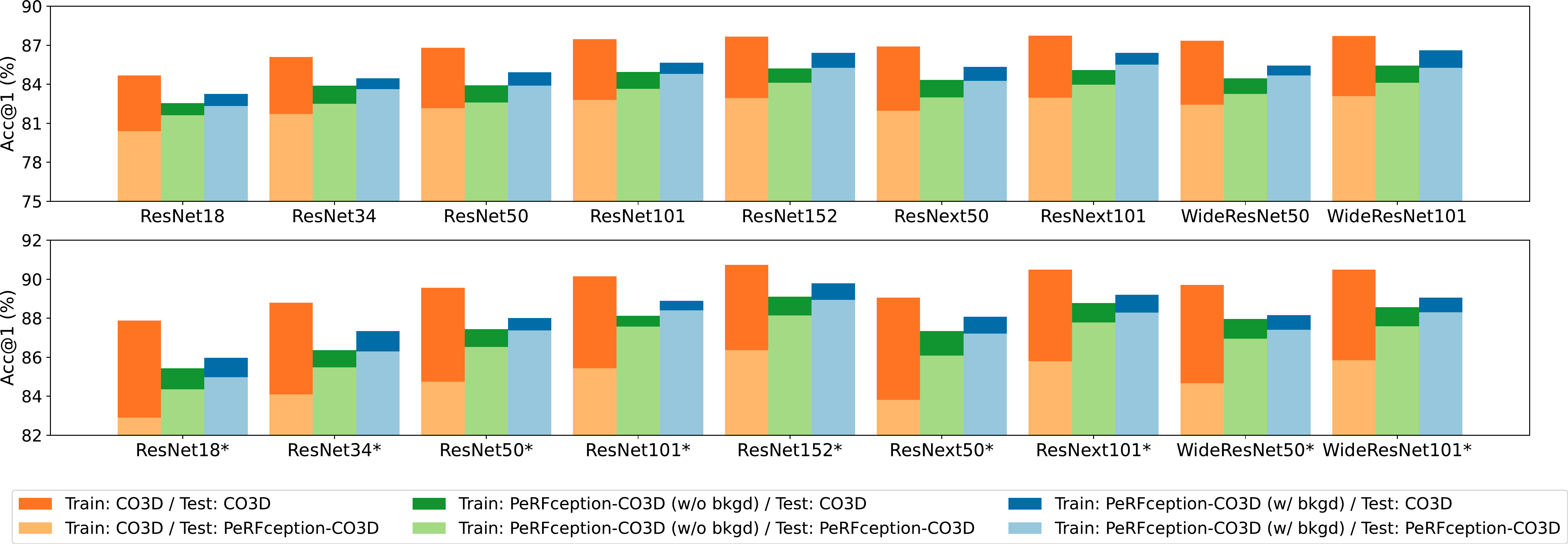}
\vspace{-4.0mm}
\caption{2D classification accuracies (Acc@1) of the ResNet models trained either on CO3D or \our-CO3D and evaluated either on CO3D or \our-CO3D. * denotes the ImageNet\cite{deng2009imagenet} pretrained network. 
The models trained on \our CO3D dataset perform well on both CO3D test dataset and \our CO3D test dataset. Furthermore, the background augmentation of \our CO3D dataset improves the 2D classification performance. The score table is in the appendix.}
\label{fig:2d_cls}
\end{figure}

We train both the scratch and ImageNet~\cite{deng2009imagenet} pretrained version of ResNet~\cite{he2016deep} variants on the original CO3D and \our{CO3D} to show that \our{CO3D} contains the same information as the original CO3D dataset. Using our random pose selection algorithm(described in the appendix), which discourages the selected pose to be extremely unobserved the train frames, we select 50 poses for each scene. As shown in Fig~\ref{fig:2d_cls}, the model trained with the original CO3D has a larger gap than the model trained with \our{CO3D} dataset. In addition, we observed using background augmentation is beneficial for improving generalizability of classification networks, especially performs the best when the augmentation is applied with probability $p=0.5$. We conduct controlled experiments about the probability $p$ in the appendix. In addition, using a popular vision analysis tool GradCAM++~\cite{chattopadhay2018grad}, we demonstrate that using background augmentation helps the model not to memorize the backgrounds in the appendix.

\subsubsection{3D classification on \our {CO3D}}

\begin{figure}[!tb]
\includegraphics[width=1.\linewidth]{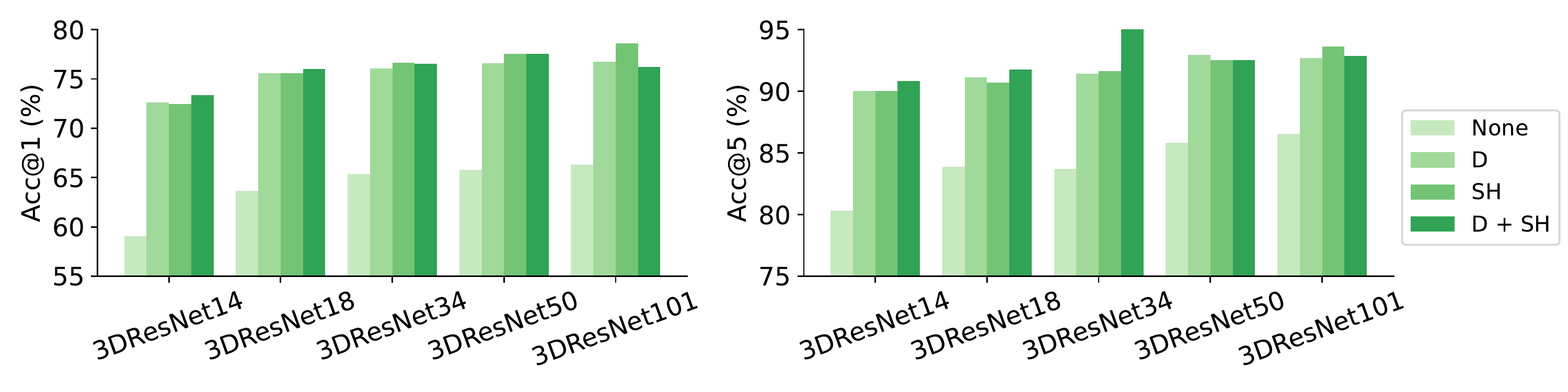}
\vspace{-4mm}
\caption{3D classification performance of 3D ResNet~\cite{choy20194d} models on our \our CO3D. We visualize Acc@1 (Left) and Acc@5 (Right) score for each model and input features. "None" denotes the case where 3D classification \textbf{D} denotes the density and \textbf{SH} denotes spherical harmonic coefficients.}
\label{fig:fig_3d_classification}
\end{figure}

\our{CO3D} provides a novel 3D data representation which we can directly feed into a network without explicit rendering. We train spatially sparse 3D networks on \our{CO3D} and visualize the 3D classification accuracy on Figure~\ref{fig:fig_3d_classification}. For each classification model, we utilize four types of input features that are in our Plenoxels representation: ones (None), densities (D), spherical harmonic coefficients (SH), and concatenation of spherical harmonic coefficients and densities (SH + D).
One interesting observation is that using either density values or spherical harmonic coefficients as features improves performance much better. We conjecture this is because the density values provide information about where the model should focus more, and the spherical harmonic coefficients explicitly encode visual features. 


\subsection{Semantic Segmentation on \our ScanNet dataset}
To further verify the fine-grained perception on the large-scale implicit data, 
we create and evaluate 3D semantic segmentation networks on our scene-centric \our ScanNet dataset.
We assign semantic labels to each voxel by aligning the reconstructed \our ScanNet data with the provided ground truth point cloud data.
Then, for each voxel of \our ScanNet data, we find the nearest point in the point cloud and when the distance is smaller than the predefined threshold (5cm for our experiments), we assign the class label of the nearest point to the voxel. Otherwise, we set the voxel label to IGNORE\_CLASS.

Similar to 3D classification experiments, we use spatially sparse convolutional networks for prediction, but we use U-shaped convnets with varying depth and width for segmentation.
For all networks, we trained for 60K iterations, with batch size 8, SGD optimizer with initial learning rate 0.1, and cosine annealing scheduler. 
We train each network with different input features as same with the 3D classification in Sec~\ref{subsec:3d_cls} to analyze the effect of the plenoptic features to the 3D semantic segmentation task.
We apply geometric augmentation (random rotation, random crop, random affine transform, coordinate dropout, random horizontal flip, random translation, elastic distortion), and feature-level augmentation (random feature jittering).

We use the standard experimental settings following ~\cite{choy20194d}, and report mean Intersection over Union (mIoU), mean per-point accuracy (mAcc) on the validation split in \Fig{3d_semseg}, and scenewise statistics are in the appendix.
We achieve up to 62.77\% mIoU and 72.36\% mAcc on the validation split of \our ScanNet, 
which shows that our \our ScanNet dataset has accurate geometry for networks to learn semantic of each object class. 
Consistent with 3D classification, the networks trained with spherical harmonic coefficients and density as input feature exhibit higher segmentation accuracy.
This results indicate that the spherical harmonic coefficients and density values provide additional cues for fine-grained geometric perception as well. 
Additional quantitative and qualitative results are provided in the appendix.

\begin{figure}[!t]
\centering
\includegraphics[width=\textwidth]{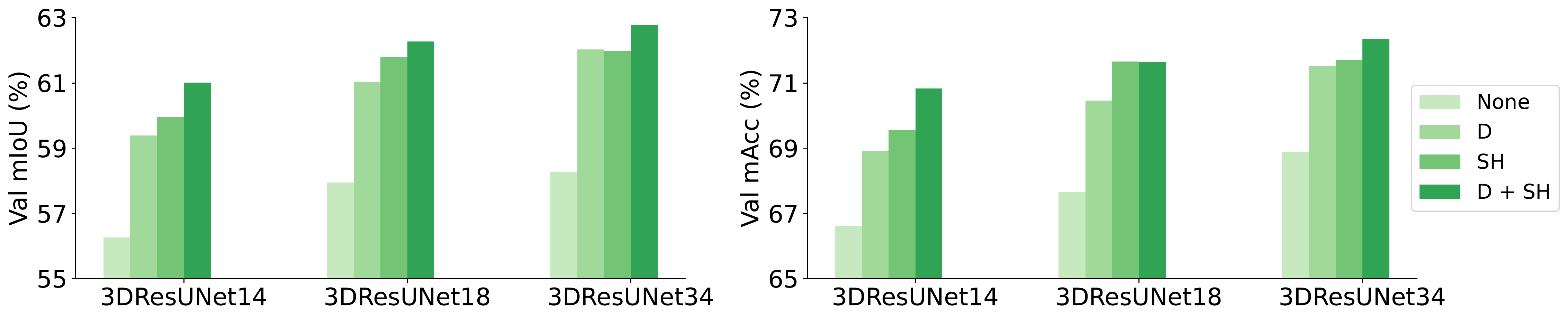}
\vspace{-4.0mm}
\caption{Evaluated semantic segmentation performance, mIoU (Left) and mAcc (Right), on \our ScanNet validation set with various input features. \textbf{D} denotes the density, \textbf{SH} denotes spherical harmonic coefficients.}
\label{fig:3d_semseg}
\end{figure}
\section{Conclusion}

In this work, we present the first perception networks for an implicit representation and conduct the comprehensive study of various visual perception tasks. 
To this end, we created two large-scale implicit datasets, namely \our CO3D and \our ScanNet, that cover object-centric and scene-centric environments, respectively.
Extensive experiments with diverse perception scenarios, including 2D image classification, 3D object classification, and 3D scene semantic segmentation, show that our datasets effectively convey the same information for both 2D and 3D in a unified and compressed data format. 
This data format allows eliminating the need to separately store different data formats, 2D images and 3D shapes. Consequently, the required disk space for storage is reduced and the unified data format includes richer features.
Furthermore, we propose a novel image augmentation method that was infeasible in image datasets. 
We expect our fully automatic pipeline should be a great candidate for establishing equally large datasets on 3D to tremendously large 2D image datasets, potentially enabling larger models to be trained. 

\revision{\textbf{Limitation.}
Plenoxels allow high-quality rendering for both indoor and outdoor scenes with fast training and rendering speed. However, the training step of Plenoxels strongly relies on calibrated camera information. The camera parameters would be inaccurately calibrated in the scenes when there are lots of symmetric or textureless patterns. Wrong camera information involves severe artifacts on rendered images, such as the occurrence of floater or geometrically deformed voxel shapes. We believe that jointly optimizing camera poses would be beneficial for improving the fidelity of our dataset. Our work opens up the potential of using radiance field representation in some conventional visual perception tasks and provides the first large-scale radiance field datasets that effectively convey both the 2D and 3D information. We expect future work relevant to more accurate and fast reconstruction could improve our work.}

\bibliographystyle{unsrt}
\bibliography{main}

\clearpage
\renewcommand{\theequation}{a.\arabic{equation}}
\renewcommand{\thetable}{a.\arabic{table}}
\renewcommand{\thefigure}{a.\arabic{figure}}
\renewcommand*{\thefootnote}{\arabic{footnote}}
\renewcommand\thesection{\Alph{section}}
\setcounter{section}{0}
\setcounter{figure}{0}
\setcounter{table}{0}
\addcontentsline{toc}{section}{Appendices}
\section*{Appendix}
\setcounter{section}{1}

\subsection{Effect of Background Augmentation}

\begin{figure*}[!htb]
\centering

\includegraphics[width=0.995\textwidth]{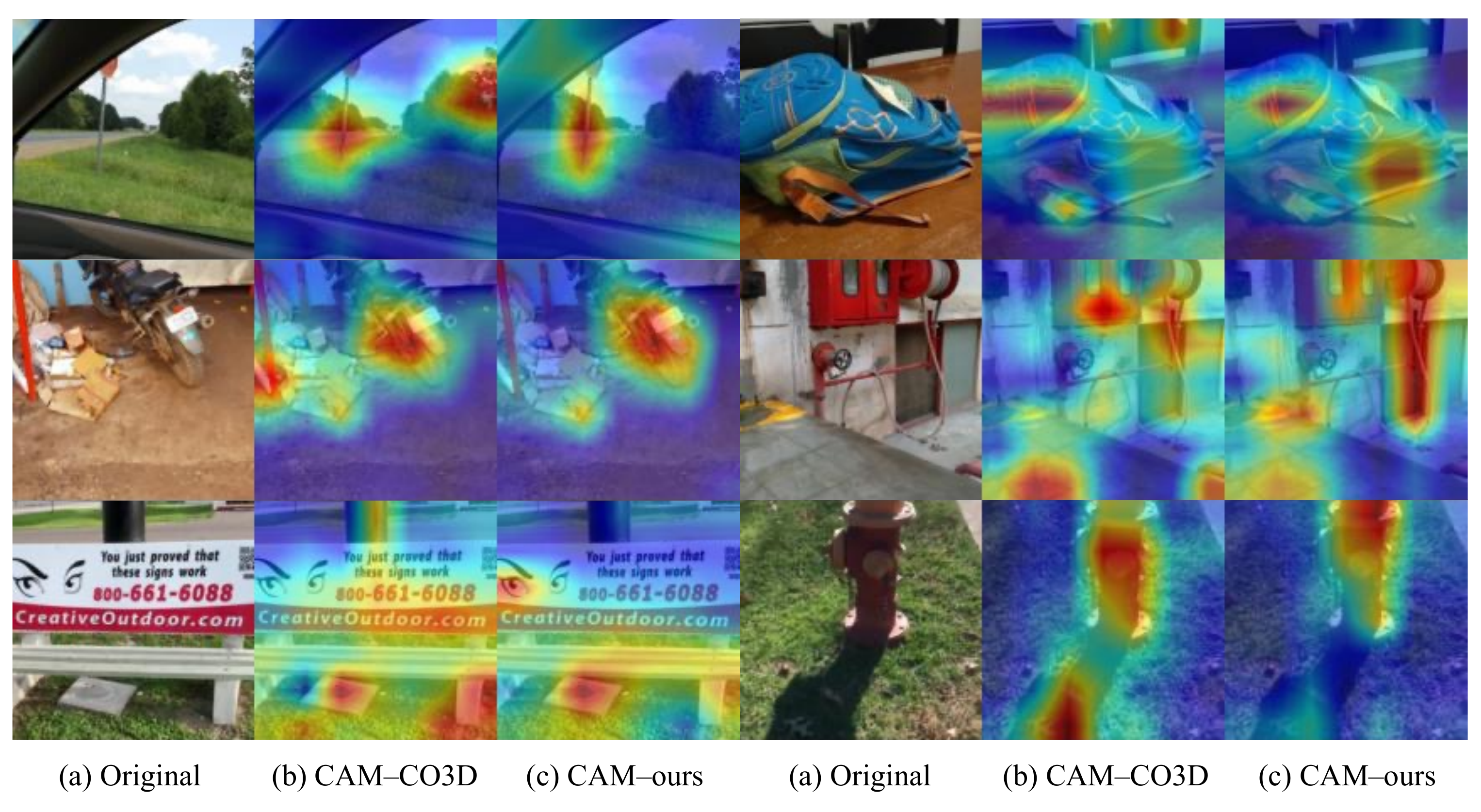}
\caption{Visualization of class activation maps generated by GradCAM++. The left and right activation map are generated from the ResNet18 trained with the original CO3D images and \our-CO3D images, respectively. }
\label{fig:qual_sup_cam}
\end{figure*}

\noindent Reducing bias on the dataset is one of the fundamental problems in computer vision. 
A typical example of data bias in object-centric dataset is the backgrounds of image that contain locational context. For instance, couch images are mostly captured in the living room, and motorcycle images are taken outdoor hence the roadway or garage are appears in the background most of the time. Such a strong correlation between background information and the class labels makes the classification networks biased toward background so that the networks make predictions by looking at the backgrounds rather than the foreground objects. Here, we demonstrate that our datasets, with additional augmentations, are more resistant to such bias.
We use a popular analysis tool, Grad-CAM++~\cite{chattopadhay2018grad}, to verify that the classification model trained with our dataset avoids overfitting on backgrounds. We use the official Grad-CAM++ implementation and feed test images. According to Figure~\ref{fig:qual_sup_cam}, the classification model trained on our dataset focuses on the object, whereas the classification model trained on the original CO3D focuses on the background for many outdoor scenes. 

We also conduct control experiments about the strength of background augmentation. In detail,  we compare evaluated performance by changing the probability, \textit{i.e,} 0\%, 50\%, and 100\%, to apply the background augmentation for each iteration. In comparison to the results without background augmentation, the background augmentation with a probability of 50\% increases the 2D classification accuracy; whereas the 100\% probability of background augmentation does not improve the performance. In this ablation study, we determine the appropriate level of background augmentation assistance for improving 2D classification accuracy. Table~\ref{tab:co3d_2d_cls} reports the results of the control experiments.

\begin{table}[!t]
\setlength{\tabcolsep}{2pt} 
\centering
\caption{2D classification accuracies (Acc@1/Acc@5) of the ResNet model trained either on CO3D or \our-CO3D and evaluated either on CO3D or \our-CO3D. * denotes the ImageNet\cite{deng2009imagenet} pretrained network. PeRF-CO3D is an abbreviation of PeRFception-CO3D. $p$ stands for the probability to apply background augmentation.}
\resizebox{.98\textwidth}{!}{
\begin{tabular}{ccccccccc}
\toprule
Train Dataset  & \multicolumn{2}{c}{CO3D} & \multicolumn{2}{c}{\revision{PeRF}-CO3D (p=0.0)} & \multicolumn{2}{c}{\revision{PeRF}-CO3D (p=0.5)} & \multicolumn{2}{c}{\revision{PeRF}-CO3D (p=1.0)}  \\
\cmidrule(lr){1-1} \cmidrule(lr){2-3} \cmidrule(lr){4-5} \cmidrule(lr){6-7} \cmidrule(lr){8-9}
\multicolumn{1}{c}{Test Dataset} & CO3D & \revision{PeRF}-CO3D & CO3D & \revision{PeRF}-CO3D & CO3D & \revision{PeRF}-CO3D & CO3D & \revision{PeRF}-CO3D  \\
\midrule
             \multicolumn{9}{c}{Acc@1 ($\mu \pm \sigma$) } \\
\midrule
ResNet18 & 84.74 $\pm$ 0.04 & \cellcolor{yellow!40}{80.39 $\pm$ 0.19} & 82.56 $\pm$ 0.24 & \cellcolor{yellow!40}{81.43 $\pm$ 0.14} & 83.15 $\pm$ 0.10 & \cellcolor{yellow!40}{82.05 $\pm$ 0.24} & 81.93 $\pm$ 0.10 & \cellcolor{yellow!40}{81.31 $\pm$ 0.12} \\
ResNet34 & 86.18 $\pm$ 0.05 & \cellcolor{yellow!40}{81.62 $\pm$ 0.10} & 83.98 $\pm$ 0.07 & \cellcolor{yellow!40}{82.71 $\pm$ 0.17} & 84.62 $\pm$ 0.14 & \cellcolor{yellow!40}{83.61 $\pm$ 0.04} & 83.40 $\pm$ 0.10 & \cellcolor{yellow!40}{82.54 $\pm$ 0.22} \\
ResNet50 & 86.83 $\pm$ 0.02 & \cellcolor{yellow!40}{82.19 $\pm$ 0.09} & 84.12 $\pm$ 0.28 & \cellcolor{yellow!40}{82.80 $\pm$ 0.37} & 84.83 $\pm$ 0.14 & \cellcolor{yellow!40}{83.77 $\pm$ 0.08} & 83.60 $\pm$ 0.11 & \cellcolor{yellow!40}{82.73 $\pm$ 0.04} \\
ResNet101 & 87.42 $\pm$ 0.04 & \cellcolor{yellow!40}{82.86 $\pm$ 0.06} & 84.97 $\pm$ 0.13 & \cellcolor{yellow!40}{83.71 $\pm$ 0.13} & 85.95 $\pm$ 0.22 & \cellcolor{yellow!40}{85.11 $\pm$ 0.23} & 85.01 $\pm$ 0.08 & \cellcolor{yellow!40}{83.82 $\pm$ 0.05} \\
ResNet152 & 87.75 $\pm$ 0.11 & \cellcolor{yellow!40}{83.04 $\pm$ 0.11} & 85.67 $\pm$ 0.31 & \cellcolor{yellow!40}{84.35 $\pm$ 0.19} & 86.40 $\pm$ 0.04 & \cellcolor{yellow!40}{85.28 $\pm$ 0.02} & 85.11 $\pm$ 0.09 & \cellcolor{yellow!40}{84.06 $\pm$ 0.09} \\
ResNext50 & 86.84 $\pm$ 0.29 & \cellcolor{yellow!40}{81.99 $\pm$ 0.14} & 84.21 $\pm$ 0.08 & \cellcolor{yellow!40}{83.04 $\pm$ 0.04} & 85.25 $\pm$ 0.16 & \cellcolor{yellow!40}{84.32 $\pm$ 0.18} & 84.06 $\pm$ 0.38 & \cellcolor{yellow!40}{83.23 $\pm$ 0.28} \\
ResNext101 & 87.73 $\pm$ 0.03 & \cellcolor{yellow!40}{82.90 $\pm$ 0.07} & 85.22 $\pm$ 0.09 & \cellcolor{yellow!40}{84.17 $\pm$ 0.15} & 86.37 $\pm$ 0.10 & \cellcolor{yellow!40}{85.48 $\pm$ 0.06} & 85.46 $\pm$ 0.16 & \cellcolor{yellow!40}{84.46 $\pm$ 0.11} \\
WideResNet50 & 87.32 $\pm$ 0.05 & \cellcolor{yellow!40}{82.45 $\pm$ 0.11} & 84.68 $\pm$ 0.15 & \cellcolor{yellow!40}{83.64 $\pm$ 0.27} & 85.58 $\pm$ 0.10 & \cellcolor{yellow!40}{84.68 $\pm$ 0.02} & 84.35 $\pm$ 0.12 & \cellcolor{yellow!40}{83.46 $\pm$ 0.24} \\
WideResNet101 & 87.80 $\pm$ 0.12 & \cellcolor{yellow!40}{83.04 $\pm$ 0.10} & 85.36 $\pm$ 0.26 & \cellcolor{yellow!40}{84.15 $\pm$ 0.19} & 86.32 $\pm$ 0.21 & \cellcolor{yellow!40}{85.30 $\pm$ 0.11} & 85.45 $\pm$ 0.13 & \cellcolor{yellow!40}{84.23 $\pm$ 0.23} \\
\midrule
ResNet18* & 87.75 $\pm$ 0.18 & \cellcolor{yellow!40}{82.70 $\pm$ 0.15} & 85.28 $\pm$ 0.14 & \cellcolor{yellow!40}{84.09 $\pm$ 0.20} & 85.97 $\pm$ 0.16 & \cellcolor{yellow!40}{84.97 $\pm$ 0.13} & 84.62 $\pm$ 0.02 & \cellcolor{yellow!40}{83.62 $\pm$ 0.13} \\
ResNet34* & 88.83 $\pm$ 0.06 & \cellcolor{yellow!40}{84.01 $\pm$ 0.12} & 86.60 $\pm$ 0.27 & \cellcolor{yellow!40}{85.43 $\pm$ 0.22} & 87.19 $\pm$ 0.31 & \cellcolor{yellow!40}{86.25 $\pm$ 0.19} & 86.09 $\pm$ 0.11 & \cellcolor{yellow!40}{85.01 $\pm$ 0.09} \\
ResNet50* & 89.51 $\pm$ 0.21 & \cellcolor{yellow!40}{84.76 $\pm$ 0.04} & 87.49 $\pm$ 0.11 & \cellcolor{yellow!40}{86.60 $\pm$ 0.08} & 88.12 $\pm$ 0.08 & \cellcolor{yellow!40}{87.30 $\pm$ 0.08} & 87.09 $\pm$ 0.12 & \cellcolor{yellow!40}{86.25 $\pm$ 0.17} \\
ResNet101* & 90.21 $\pm$ 0.09 & \cellcolor{yellow!40}{85.60 $\pm$ 0.16} & 88.39 $\pm$ 0.22 & \cellcolor{yellow!40}{87.46 $\pm$ 0.17} & 89.00 $\pm$ 0.07 & \cellcolor{yellow!40}{88.32 $\pm$ 0.13} & 88.28 $\pm$ 0.07 & \cellcolor{yellow!40}{87.26 $\pm$ 0.01} \\
ResNet152* & 90.60 $\pm$ 0.10 & \cellcolor{yellow!40}{86.26 $\pm$ 0.10} & 89.17 $\pm$ 0.15 & \cellcolor{yellow!40}{88.19 $\pm$ 0.03} & 89.52 $\pm$ 0.20 & \cellcolor{yellow!40}{88.73 $\pm$ 0.15} & 88.63 $\pm$ 0.20 & \cellcolor{yellow!40}{87.59 $\pm$ 0.13} \\
ResNext50* & 89.21 $\pm$ 0.14 & \cellcolor{yellow!40}{83.90 $\pm$ 0.28} & 87.28 $\pm$ 0.06 & \cellcolor{yellow!40}{86.28 $\pm$ 0.14} & 87.82 $\pm$ 0.19 & \cellcolor{yellow!40}{87.30 $\pm$ 0.06} & 86.81 $\pm$ 0.27 & \cellcolor{yellow!40}{86.09 $\pm$ 0.25} \\
ResNext101* & 90.52 $\pm$ 0.07 & \cellcolor{yellow!40}{85.91 $\pm$ 0.13} & 88.51 $\pm$ 0.19 & \cellcolor{yellow!40}{87.82 $\pm$ 0.06} & 89.17 $\pm$ 0.07 & \cellcolor{yellow!40}{88.51 $\pm$ 0.16} & 88.57 $\pm$ 0.02 & \cellcolor{yellow!40}{87.60 $\pm$ 0.18} \\
WideResNet50* & 89.78 $\pm$ 0.06 & \cellcolor{yellow!40}{85.13 $\pm$ 0.33} & 87.80 $\pm$ 0.18 & \cellcolor{yellow!40}{86.88 $\pm$ 0.06} & 88.23 $\pm$ 0.10 & \cellcolor{yellow!40}{87.75 $\pm$ 0.25} & 87.12 $\pm$ 0.11 & \cellcolor{yellow!40}{86.50 $\pm$ 0.08} \\
WideResNet101* & 90.45 $\pm$ 0.10 & \cellcolor{yellow!40}{85.97 $\pm$ 0.09} & 88.53 $\pm$ 0.04 & \cellcolor{yellow!40}{87.59 $\pm$ 0.12} & 89.22 $\pm$ 0.13 & \cellcolor{yellow!40}{88.39 $\pm$ 0.07} & 88.10 $\pm$ 0.11 & \cellcolor{yellow!40}{87.20 $\pm$ 0.10} \\
\midrule
             \multicolumn{9}{c}{Acc@5 ($\mu \pm \sigma$) } \\
\midrule
ResNet18 & 96.24 $\pm$ 0.17 & \cellcolor{yellow!40}{94.19 $\pm$ 0.16} & 95.55 $\pm$ 0.18 & \cellcolor{yellow!40}{95.00 $\pm$ 0.15} & 95.70 $\pm$ 0.11 & \cellcolor{yellow!40}{95.37 $\pm$ 0.05} & 95.30 $\pm$ 0.08 & \cellcolor{yellow!40}{94.85 $\pm$ 0.09} \\
ResNet34 & 96.68 $\pm$ 0.11 & \cellcolor{yellow!40}{94.62 $\pm$ 0.11} & 95.88 $\pm$ 0.07 & \cellcolor{yellow!40}{95.37 $\pm$ 0.10} & 96.23 $\pm$ 0.03 & \cellcolor{yellow!40}{95.89 $\pm$ 0.06} & 95.74 $\pm$ 0.11 & \cellcolor{yellow!40}{95.29 $\pm$ 0.07} \\
ResNet50 & 96.90 $\pm$ 0.07 & \cellcolor{yellow!40}{94.67 $\pm$ 0.14} & 96.10 $\pm$ 0.02 & \cellcolor{yellow!40}{95.51 $\pm$ 0.13} & 96.41 $\pm$ 0.11 & \cellcolor{yellow!40}{95.99 $\pm$ 0.08} & 96.01 $\pm$ 0.07 & \cellcolor{yellow!40}{95.48 $\pm$ 0.04} \\
ResNet101 & 97.04 $\pm$ 0.03 & \cellcolor{yellow!40}{94.78 $\pm$ 0.05} & 96.23 $\pm$ 0.07 & \cellcolor{yellow!40}{95.58 $\pm$ 0.09} & 96.76 $\pm$ 0.10 & \cellcolor{yellow!40}{96.32 $\pm$ 0.12} & 96.39 $\pm$ 0.03 & \cellcolor{yellow!40}{95.74 $\pm$ 0.14} \\
ResNet152 & 97.14 $\pm$ 0.04 & \cellcolor{yellow!40}{94.94 $\pm$ 0.09} & 96.32 $\pm$ 0.09 & \cellcolor{yellow!40}{95.73 $\pm$ 0.05} & 96.86 $\pm$ 0.06 & \cellcolor{yellow!40}{96.39 $\pm$ 0.06} & 96.46 $\pm$ 0.07 & \cellcolor{yellow!40}{95.92 $\pm$ 0.09} \\
ResNext50 & 96.75 $\pm$ 0.12 & \cellcolor{yellow!40}{94.54 $\pm$ 0.10} & 95.95 $\pm$ 0.15 & \cellcolor{yellow!40}{95.32 $\pm$ 0.13} & 96.49 $\pm$ 0.10 & \cellcolor{yellow!40}{95.92 $\pm$ 0.16} & 96.08 $\pm$ 0.12 & \cellcolor{yellow!40}{95.47 $\pm$ 0.15} \\
ResNext101 & 96.98 $\pm$ 0.07 & \cellcolor{yellow!40}{94.71 $\pm$ 0.12} & 96.09 $\pm$ 0.11 & \cellcolor{yellow!40}{95.54 $\pm$ 0.07} & 96.71 $\pm$ 0.01 & \cellcolor{yellow!40}{96.26 $\pm$ 0.03} & 96.49 $\pm$ 0.04 & \cellcolor{yellow!40}{95.80 $\pm$ 0.03} \\
WideResNet50 & 96.84 $\pm$ 0.13 & \cellcolor{yellow!40}{94.60 $\pm$ 0.10} & 96.05 $\pm$ 0.04 & \cellcolor{yellow!40}{95.52 $\pm$ 0.05} & 96.44 $\pm$ 0.11 & \cellcolor{yellow!40}{96.03 $\pm$ 0.03} & 96.10 $\pm$ 0.16 & \cellcolor{yellow!40}{95.55 $\pm$ 0.01} \\
WideResNet101 & 97.05 $\pm$ 0.01 & \cellcolor{yellow!40}{94.91 $\pm$ 0.08} & 96.26 $\pm$ 0.07 & \cellcolor{yellow!40}{95.70 $\pm$ 0.08} & 96.86 $\pm$ 0.11 & \cellcolor{yellow!40}{96.31 $\pm$ 0.10} & 96.49 $\pm$ 0.06 & \cellcolor{yellow!40}{95.89 $\pm$ 0.01} \\
\midrule
ResNet18* & 97.13 $\pm$ 0.10 & \cellcolor{yellow!40}{95.04 $\pm$ 0.16} & 96.39 $\pm$ 0.08 & \cellcolor{yellow!40}{95.89 $\pm$ 0.03} & 96.73 $\pm$ 0.04 & \cellcolor{yellow!40}{96.24 $\pm$ 0.09} & 96.19 $\pm$ 0.08 & \cellcolor{yellow!40}{95.67 $\pm$ 0.10} \\
ResNet34* & 97.39 $\pm$ 0.14 & \cellcolor{yellow!40}{95.28 $\pm$ 0.15} & 96.67 $\pm$ 0.08 & \cellcolor{yellow!40}{96.18 $\pm$ 0.04} & 97.00 $\pm$ 0.13 & \cellcolor{yellow!40}{96.50 $\pm$ 0.09} & 96.61 $\pm$ 0.05 & \cellcolor{yellow!40}{96.02 $\pm$ 0.03} \\
ResNet50* & 97.60 $\pm$ 0.04 & \cellcolor{yellow!40}{95.46 $\pm$ 0.02} & 96.84 $\pm$ 0.04 & \cellcolor{yellow!40}{96.39 $\pm$ 0.07} & 97.11 $\pm$ 0.05 & \cellcolor{yellow!40}{96.69 $\pm$ 0.08} & 96.95 $\pm$ 0.06 & \cellcolor{yellow!40}{96.37 $\pm$ 0.13} \\
ResNet101* & 97.90 $\pm$ 0.03 & \cellcolor{yellow!40}{95.86 $\pm$ 0.05} & 97.06 $\pm$ 0.19 & \cellcolor{yellow!40}{96.74 $\pm$ 0.11} & 97.48 $\pm$ 0.02 & \cellcolor{yellow!40}{97.13 $\pm$ 0.05} & 97.16 $\pm$ 0.04 & \cellcolor{yellow!40}{96.57 $\pm$ 0.03} \\
ResNet152* & 97.97 $\pm$ 0.04 & \cellcolor{yellow!40}{95.97 $\pm$ 0.04} & 97.32 $\pm$ 0.17 & \cellcolor{yellow!40}{96.87 $\pm$ 0.13} & 97.66 $\pm$ 0.07 & \cellcolor{yellow!40}{97.24 $\pm$ 0.08} & 97.22 $\pm$ 0.17 & \cellcolor{yellow!40}{96.69 $\pm$ 0.09} \\
ResNext50* & 97.33 $\pm$ 0.09 & \cellcolor{yellow!40}{94.99 $\pm$ 0.11} & 96.54 $\pm$ 0.09 & \cellcolor{yellow!40}{96.04 $\pm$ 0.02} & 97.08 $\pm$ 0.14 & \cellcolor{yellow!40}{96.66 $\pm$ 0.07} & 96.57 $\pm$ 0.12 & \cellcolor{yellow!40}{96.03 $\pm$ 0.04} \\
ResNext101* & 97.67 $\pm$ 0.03 & \cellcolor{yellow!40}{95.51 $\pm$ 0.06} & 96.90 $\pm$ 0.02 & \cellcolor{yellow!40}{96.61 $\pm$ 0.03} & 97.32 $\pm$ 0.05 & \cellcolor{yellow!40}{96.93 $\pm$ 0.07} & 97.07 $\pm$ 0.04 & \cellcolor{yellow!40}{96.52 $\pm$ 0.05} \\
WideResNet50* & 97.60 $\pm$ 0.07 & \cellcolor{yellow!40}{95.53 $\pm$ 0.17} & 96.75 $\pm$ 0.10 & \cellcolor{yellow!40}{96.39 $\pm$ 0.08} & 97.13 $\pm$ 0.02 & \cellcolor{yellow!40}{96.84 $\pm$ 0.09} & 96.72 $\pm$ 0.07 & \cellcolor{yellow!40}{96.22 $\pm$ 0.05} \\
WideResNet101* & 97.76 $\pm$ 0.07 & \cellcolor{yellow!40}{95.67 $\pm$ 0.04} & 97.01 $\pm$ 0.05 & \cellcolor{yellow!40}{96.62 $\pm$ 0.07} & 97.48 $\pm$ 0.07 & \cellcolor{yellow!40}{97.10 $\pm$ 0.06} & 96.93 $\pm$ 0.09 & \cellcolor{yellow!40}{96.48 $\pm$ 0.06} \\
\bottomrule
\end{tabular}
}
\label{tab:co3d_2d_cls}
\end{table}

We also visualize several examples of background augmented images on Figure~\ref{fig:augmentation}.
\subsection{Implementation Details}

Here, we describe the thorough details about the data generation with Plenoxels and the data augmentation methods proposed in our main paper.
We also provide the architectural details of 3D semantic segmentation models trained on \our ScanNet dataset.

\subsubsection{Data Generation}
We use the official implementation of Plenoxels~\cite{yu2021plenoxels}, which is implemented in PyTorch with custom CUDA kernels. With a single RTX 3090 GPU, it takes up to 30 minutes per scene. 

\textbf{\our CO3D.}
Plenoxels uses a different learning rate for each parameter. Following the default configuration for rendering Tanks and Temples~\cite{knapitsch2017tanks}
dataset, we use the learning rate 30 for density values and $10^{-2}$ for spherical harmonic coefficients. We skip foreground rendering for 1,000 steps at the beginning of training for stable training. For total variation losses, we set the weight $\lambda$ as $5.0 \times 10^{-5}$ for foreground densities, $5.0 \times 10^{-3}$ for foreground spherical harmonics, $1.0 \times 10^{-3}$ for background colors, and $1.0 \times 10^{-3}$ for background densities. The background brightness is set to 0.5. We set the lambda value to $10^{-5}$ for the beta loss and $10^{-10}$ for the sparsity loss, which is 10 times larger than the default configuration. We use the 9-dimensional spherical harmonic coefficients for each RGB channel. We prune voxels whose density values are below 1.28 during the upsampling step. 

\textbf{\our ScanNet.}
As we described in Section 4.2 in our main paper, we initialize the voxel grid with unprojected depth maps for \our ScanNet dataset generation.
The unprojected depth maps are aggregated into a point cloud using the ground truth camera parameters. 
Since the ScanNet depth maps often contain noisy observations, the resulting point cloud is contaminated with outlier points. 
To filter out the outliers, we discretize the point cloud into coarse voxels with discretization size of 5cm and perform connected component analysis to detect disconnected voxels which are likely to be outliers.
The refined point cloud is then utilized as the initialization for Plenoxels.
The voxel grid is initialized with 256$^3$ resolution and trained for 51,200 iterations and at 25,600th iteration we perform voxel pruning with the density threshold of 1.28.
All the other hyperparameters are same with CO3D but removed background rendering since all the points of ScanNet datasets should be considered as foregrounds.

The input images are resized into 640*480 resolution. We exclude 16 scenes out of 1,513 training scenes from ScanNet that are severely affected by motion blurs. Detecting blurriness was done automatically by OpenCV blur detection algorithm with the threshold of 10. We consider a scene defective when the number of frames is too low after filtering blur images. 

\subsubsection{Random Pose Selection}

Plenoxels show great rendering quality when the camera pose is close to the camera poses in the train set. 
Unfortunately, it fails to reliably render frames when attempting to render extremely unobserved parts, which are significantly out of train-view coverage.
Thus, selecting appropriate poses is crucial for photorealistic image generation.
We propose an operation that selects an intermediate pose from two input poses. The operator $\operatorname{AA2R}$ converts the axis-angle representation, i.e., $(v, \theta)$ where $v$ stands for the axis and $\theta$ stands for the angle,  to the rotation matrix. $\operatorname{R2AA}$ is exactly the inverse operation of $\operatorname{AA2R}$. We compute the distance between the selected poses with distance functions $D_R$ and $D_t$ to choose a pair of poses that are close enough. We use the rotation threshold $\theta = \frac{1}{24}\pi$ and the translation threshold $0.5$. 

\begin{equation*}
    \operatorname{ReduceAngle}(\mathbf{R}, s) = \operatorname{AA2R}(\operatorname{R2AA}(\mathbf{R})_{v}, s \cdot \operatorname{R2AA}(\mathbf{R})_{\theta})
\end{equation*}
\begin{equation*}
    \operatorname{IntermediateRot}(\mathbf{R}_1, \mathbf{R}_2, s) = \operatorname{ReduceAngle}(\mathbf{R}_2\mathbf{R}_1^{-1}, s) \mathbf{R}_1 
\end{equation*}
\begin{equation*}
    D_R(\mathbf{R}_1, \mathbf{R}_2) = \arccos{(\mathrm{Tr}(\mathbf{R}_2^{-1}\mathbf{R}_1) / 2 - 1)}
\end{equation*}
\begin{equation*}
    D_t(\mathbf{t}_1, \mathbf{t}_2) = ||\mathbf{t}_2 - \mathbf{t}_1||_2^2
\end{equation*}

\noindent The main concept of the Algo~\ref{alg:pose_aug} is to select an intermediate camera pose between randomly selected camera poses in the train set. 
For intrinsic parameters and image shape, we randomly select intrinsic matrix among intrinsic matrices in train set and use the corresponding image shape.

\begin{algorithm}
    \caption{Random Pose Generation}
    \begin{algorithmic}        
        \State{Input}
            \State{\quad $P = \{ \{ \mathbf{R}_i, \mathbf{t}_i \}\}_{i = 1, 2, \cdots N}$ : The set of poses in train split}
            \State{\quad $\theta$: The rotation distance threshold}   
            \State{\quad $d$: The translation distance threshold} 
        \State{Output}
            \State{\quad $\mathbf{p}_{\mathrm{out}} \in \mathbb{R}^{4 \times 4}$ : The output pose}\\
        
        \State{$s$: A random value from $[0, 1]$}
        \State{$j, k = \operatorname{RandomIndex}(N), \operatorname{RandomIndex}(N) $} 
        \While{$D_{R}(\textbf{R}_j, \textbf{R}_k) \geq \theta$ and $D_{t}(\textbf{t}_j, \textbf{t}_k) \geq d$} 
            \State{$j, k = \operatorname{RandomIndex}(N), \operatorname{RandomIndex}(N) $} 
        \EndWhile
        \State{$\mathbf{R}_{jk} = \operatorname{IntermediateRot}(\mathbf{R}_j, \mathbf{R}_k, s)$}
        \State{$\mathbf{p}_{\mathrm{out}}$[:3, :3]$ = \mathbf{R}_{jk}$}
        \State{$\mathbf{p}_{\mathrm{out}}[$:3$, 3] = s \mathbf{t}_k + (1 - s) \mathbf{t}_j$}
        \State{$\mathbf{p}_{\mathrm{out}}[3, 3] = 1$} \\
        \Return{$\mathbf{p}_{\mathrm{out}}$}        
    \end{algorithmic}
    \label{alg:pose_aug}
\end{algorithm}

\subsubsection{Architectural Details}

We train three variants of 3D ResUNets implemented with sparse convolutional layers~\cite{choy20194d} for semantic segmentation on \our ScanNet dataset. 
The layer-wise architectural details are depicted in \Tbl{architecture}.
\begin{table*}[]
\centering
\small
\caption{Architectures of Res16UNet variants for semantic segmentation on \our ScanNet. We denote a convolution layer with its \emph{kernel size}, \emph{output channel size}, and \emph{convolution stride size}. All convolution layers except for the last layer have a Batch Normalization and a ReLU layer after them. The layers with the tag "conv\_tr" indicates the transposed convolution layers. We use a square bracket to denote a residual block, with the number of blocks stacked.}
\vspace{2mm}
  \resizebox{0.80\textwidth}{!}{
  \begin{tabular}{l|c|c|c}
    \toprule
    layer name & Res16UNet14A & Res16UNet18A & Res16UNet34C \\ 
    \midrule
    init & \multicolumn{3}{c}{$3^3$, 32, stride 1} \\
    \midrule
    conv1 & \multicolumn{3}{c}{$2^3$, 32, stride 2} \\
    \midrule
    block1 & $\begin{bmatrix} 3^3, 32, \text{stride 1} \\ 3^3, 32, \text{stride 1}  \end{bmatrix} \times 1$ & $\begin{bmatrix} 3^3, \text{32}, \text{stride 1} \\ 3^3, 32, \text{stride 1} \end{bmatrix} \times 2$ & $\begin{bmatrix} 3^3, \text{32}, \text{stride 1} \\ 3^3, 32, \text{stride 1} \end{bmatrix} \times 2$ \\
    \midrule
    conv2 & \multicolumn{3}{c}{$2^3$, 32, stride 2} \\
    \midrule
    block2 & $\begin{bmatrix} 3^3, \text{64}, \text{stride 1} \\ 3^3, \text{64}, \text{stride 1} \end{bmatrix} \times 1$ & $\begin{bmatrix} 3^3, \text{64}, \text{stride 1} \\ 3^3, \text{64}, \text{stride 1} \end{bmatrix} \times 2$ & $\begin{bmatrix} 3^3, \text{64}, \text{stride 1} \\ 3^3, \text{64}, \text{stride 1} \end{bmatrix} \times 3$ \\
    \midrule
    conv3 & \multicolumn{3}{c}{$2^3$, 64, stride 2} \\
    \midrule
    block3 & $\begin{bmatrix} 3^3, \text{128}, \text{stride 1} \\ 3^3, \text{128}, \text{stride 1} \end{bmatrix} \times 1$ & $\begin{bmatrix} 3^3, \text{128}, \text{stride 1} \\ 3^3, \text{128}, \text{stride 1} \end{bmatrix} \times 2$ & $\begin{bmatrix} 3^3, \text{128}, \text{stride 1} \\ 3^3, \text{128}, \text{stride 1} \end{bmatrix} \times 4$ \\
    \midrule
    conv4 & \multicolumn{3}{c}{$2^2$, 128, stride 2} \\
    \midrule
    block4 & $\begin{bmatrix} 3^3, \text{256}, \text{stride 1} \\ 3^3, \text{256}, \text{stride 1} \end{bmatrix} \times 1$ & $\begin{bmatrix} 3^3, \text{256}, \text{stride 1} \\ 3^3, \text{256}, \text{stride 1} \end{bmatrix} \times 2$ & $\begin{bmatrix} 3^3, \text{256}, \text{stride 1} \\ 3^3, \text{256}, \text{stride 1} \end{bmatrix} \times 6$ \\
    \midrule
    conv4\_tr & \multicolumn{2}{c|}{$2^3$, 128, stride 2} & $2^3$, 256, stride 2 \\
    \midrule
    block5 & $\begin{bmatrix} 3^3, \text{128}, \text{stride 1} \\ 3^3, \text{128}, \text{stride 1} \end{bmatrix} \times 1$ & $\begin{bmatrix} 3^3, \text{128}, \text{stride 1} \\ 3^3, \text{128}, \text{stride 1} \end{bmatrix} \times 2$ & $\begin{bmatrix} 3^3, \text{256}, \text{stride 1} \\ 3^3, \text{256}, \text{stride 1} \end{bmatrix} \times 2$ \\
    \midrule
    conv5\_tr & \multicolumn{3}{c}{$2^2$, 128, stride 2} \\
    \midrule
    block6 & $\begin{bmatrix} 3^3, \text{128}, \text{stride 1} \\ 3^3, \text{128}, \text{stride 1} \end{bmatrix} \times 1$ & $\begin{bmatrix} 3^3, \text{128}, \text{stride 1} \\ 3^3, \text{128}, \text{stride 1} \end{bmatrix} \times 2$ & $\begin{bmatrix} 3^3, \text{128}, \text{stride 1} \\ 3^3, \text{128}, \text{stride 1} \end{bmatrix} \times 2$ \\
    \midrule
    conv6\_tr & \multicolumn{3}{c}{$2^3$, 96, stride 2} \\
    \midrule
    block7 & $\begin{bmatrix} 3^3, \text{96}, \text{stride 1} \\ 3^3, \text{96}, \text{stride 1} \end{bmatrix} \times 1$ & $\begin{bmatrix} 3^3, \text{96}, \text{stride 1} \\ 3^3, \text{96}, \text{stride 1} \end{bmatrix} \times 2$ & $\begin{bmatrix} 3^3, \text{96}, \text{stride 1} \\ 3^3, \text{96}, \text{stride 1} \end{bmatrix} \times 2$ \\
    \midrule
    conv7\_tr & \multicolumn{3}{c}{$2^3$, 96, stride 2} \\
    \midrule
    block8 & $\begin{bmatrix} 3^3, \text{96}, \text{stride 1} \\ 3^3, \text{96}, \text{stride 1} \end{bmatrix} \times 1$ & $\begin{bmatrix} 3^3, \text{96}, \text{stride 1} \\ 3^3, \text{96}, \text{stride 1} \end{bmatrix} \times 2$ & $\begin{bmatrix} 3^3, \text{96}, \text{stride 1} \\ 3^3, \text{96}, \text{stride 1} \end{bmatrix} \times 2$ \\
    \midrule
    final & \multicolumn{3}{c}{$1^3, 20, \text{stride 1}$} \\
    \bottomrule
  \end{tabular}
  }
\label{tbl:architecture}
\end{table*}

\subsection{Dataset Statistics}

\begin{figure*}[!htb]
\centering
\includegraphics[width=0.995\textwidth]{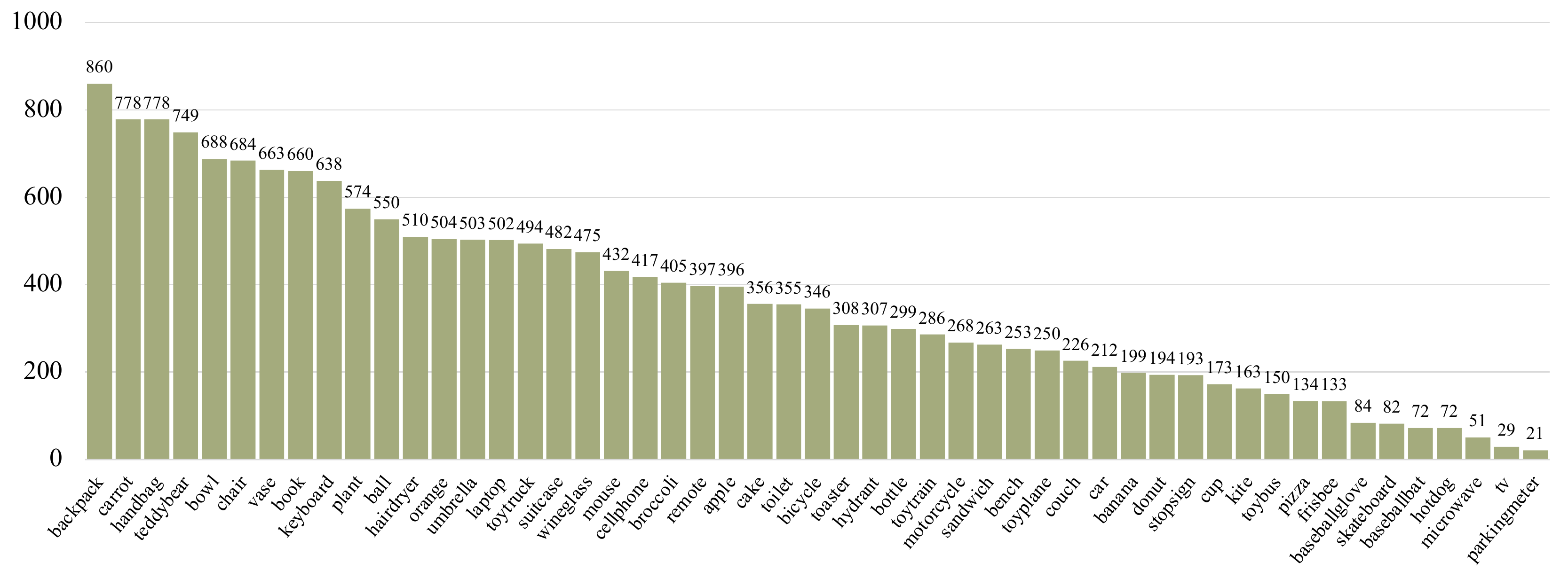}
\caption{Statistics of \our CO3D dataset scene per object-class. The y-axis visualizes the number of scenes for each class. 
}
\label{fig:dataspec}
\end{figure*}

\our CO3D dataset includes 18,618 object-centric scenes that cover all of the original CO3D dataset scenes, only except those with incorrect camera poses.
It contains a fair quantity of 51 different object-class labeled scenes; see Figure~\ref{fig:dataspec}.

\revision{
We also analyze the \textit{resolution vs. rendering quality} trade-off on our PeRFception-ScanNet dataset. To measure the trade-off rates, we train Plenoxels with lower (128) and higher (512) resolutions than the default configuration (256) on a randomly selected subset of ScanNet scenes.}

\begin{figure*}[!htb]
\centering
\includegraphics[width=0.6\textwidth]{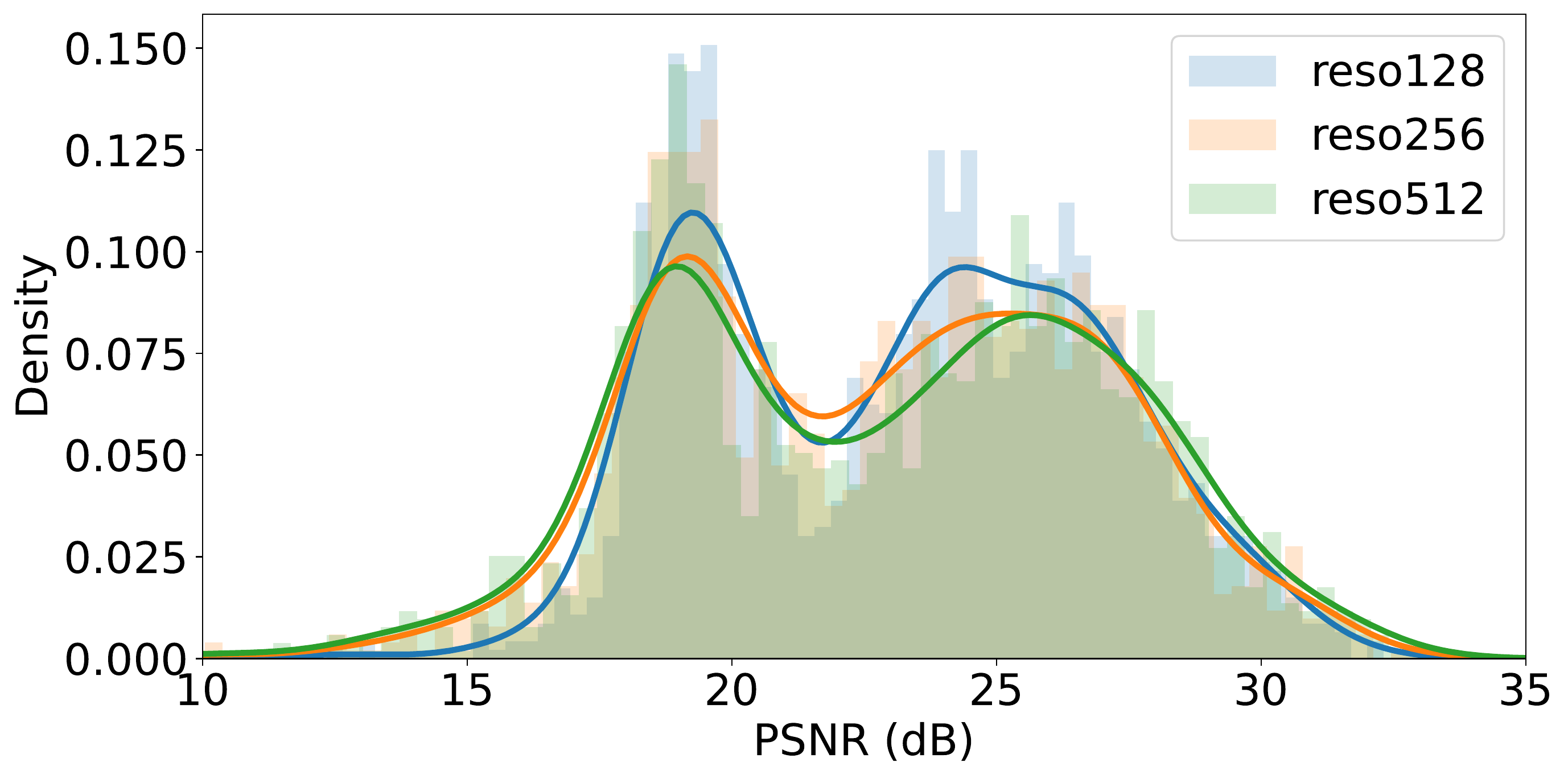}
\caption{\revision{Distributions of PSNR values of \our{}ScanNet dataset with varying resolutions.}}
\label{fig:psnr}
\end{figure*}

\revision{
As reported in \Tbl{psnr} and \Fig{psnr}, there is no direct correlation between the resolution and the average rendering quality on ScanNet reconstruction. This could be attributed to various non-trivial factors. We visualized the histogram of PSNR distribution on the ScanNet dataset in \Fig{psnr}. The X-axis represents the PSNR score, and Y-axis represents the percentage of scenes. Note that the PSNR distribution of the higher resolution Plenoxel reconstructions exhibits fat-tailed distribution, whereas the lower resolution reconstructions show the long-tailed distribution. We speculate that this is due to the fact that higher resolution reconstruction results in each voxel learning spherical harmonics parameters from fewer rays. Thus, errors in camera parameters or motion blur would result in larger errors for smaller voxels as parameters are learned from fewer rays. Thus, the rendering quality increases for the scenes with accurate camera poses and less motion blur, while it decreases for noisy scenes. We conjecture that higher resolution reconstruction would yield better performance if we have high-resolution images with accurate camera poses.
}

\begin{table}[!tb]
    \caption{\revision{PSNR of PeRFception-ScanNet dataset with varying resolutions. PCTL stands for \textit{percentile}.}}
    \centering
    \begin{tabular}{ccccccc}
        \toprule
        Reso & Mem (GB) & Avg. PSNR (dB) & 50th PCTL & 75th PCTL & 90th PCTL & 95th PCTL\\
        \midrule
        128 & 28.7 & 23.20 $\pm{3.69}$& 23.59 & 26.14 & 27.94 & 29.12 \\
        256 & 43.8 & 23.01 $\pm{3.96}$ & 23.38 & 26.10 & 28.02 & 29.20 \\
        512 & 113.8 & 22.94 $\pm{4.26}$ & 23.22 & 26.35 & 28.34 & 29.49 \\
        \bottomrule
    \end{tabular}
    \label{tbl:psnr}
\end{table}

\subsection{Camera Manipulation}

\begin{figure*}[!tb]
\centering
\includegraphics[width=0.995\textwidth]{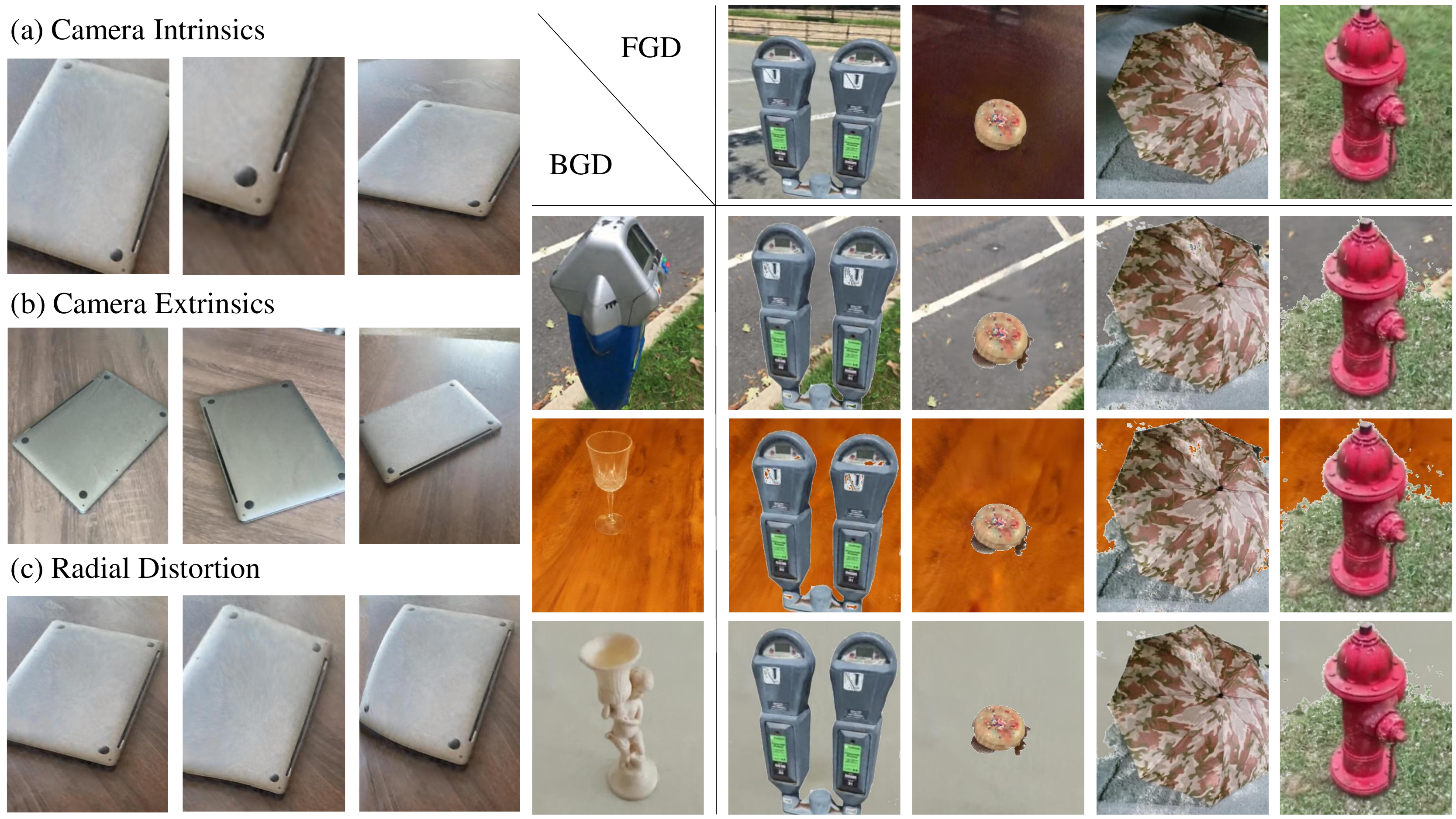}
\caption{Examples of camera manipulation and background augmentations. FGD and BGD are source images for foreground and background respectively.}
\label{fig:augmentation}
\end{figure*}
Camera manipulation has been one of the inaccessible augmentation techniques on conventional image datasets. In contrast, our implicit data can be rendered from many continuous viewpoints and with arbitrary camera parameters (\textit{e.g.}, intrinsics, extrinsics, and radial distortions), allowing us to generate images with non-standard but realistic camera-level augmentations. We have not adopted this augmentation skill while training 2D classification networks since most images do not have distortions. We perform camera intrinsics augmentation and camera distortion augmentation and Figure~\ref{fig:augmentation} visualizes several examples of such camera manipulations. 
\subsection{Quantitative Results}


In this section, we provide additional quantitative results that are not included in the main paper.

\textbf{Classwise Rendering Quality.}
We provide classwise PSNR, SSIM, and LPIPS scores of \our CO3D dataset in Table~\ref{tab:supp_classwise}.

\textbf{2D and 3D Perception Tasks.}
Table~\ref{tab:co3d_2d_cls} reports the results of the 2D classification experiments.
In Table~\ref{tab:co3d_3d_cls}, we report the quantitative classification results on \our CO3D dataset.
In \Tbl{scannet_scenewise}, we report the scenewise semantic segmentation results on \our ScanNet dataset.
For all experiments, we perform three experiments with different random seed values and report the mean and standard deviation of evaluation metrics.

\begin{table}[!htb]
\centering
\small
\resizebox{1.0\textwidth}{!}{
\begin{tabular}{cccccccccccccc}
      & apple & backp & ball & banana & bbb & bbg & bench & bicy & book & bottle & bowl & brocc & cake \\ 
\hline
PSNR  & 29.56 & 28.99 & 29.06 & 30.62 & 29.21 & 30.03 & 27.01 & 27.67 & 28.88 & 29,85 & 31.83 & 29.67 & 29.02\\
SSIM  & 0.8767 & 0.8702 & 0.8661 & 0.8951 & 0.8691 & 0.8898 & 0.7883 & 0.8246 & 0.8746 & 0.8671 & 0.8717 & 0.8584 & 0.8695 \\
LPIPS & 0.3263 & 0.3349 & 0.3614 & 0.2591 & 0.3566 & 0.3347 & 0.3471 & 0.3416 & 0.3453 & 0.3110 & 0.3135 & 0.3479 & 0.3452 \\
\end{tabular} 
}
\resizebox{1.0\textwidth}{!}{
\begin{tabular}{cccccccccccccc}
      & car & carrot & cellp & chair & couch & cup & donut & frisbee & hairdryer & handbag & hotdog & hydrant & kbd \\ 
\hline
PSNR  & 23.63 & 28.78 & 29.33 & 28.32 & 29.31 & 28.83 & 28.78 & 30.06 & 29.12 & 28.39 & 28.95 & 26.12 & 29.74\\
SSIM  & 0.7259 & 0.8573 & 0.8603 & 0.8524 & 0.8727 & 0.8628 & 0.8695 & 0.8720 & 0.8706 & 0.8622 & 0.8738 & 0.7428 & 0.8789 \\
LPIPS & 0.4003 & 0.3609 & 0.3511 & 0.3475 & 0.3358 & 0.3602 & 0.3484 & 0.3386 & 0.3351 & 0.3367 & 0.3639 & 0.3416 & 0.3301 \\
\end{tabular}
}
\resizebox{1.0\textwidth}{!}{
\begin{tabular}{cccccccccccccc}
      & kite & laptop & micro & motor & mouse & orange & park & pizza & plant & remote & sandwc & skb & stop \\ 
\hline
PSNR  & 29.04 & 27.99 & 27.75 & 25.61 & 29.04 & 28.55 & 24.69 & 29.00 & 28.63 & 29.23 & 29.00 & 28.57 & 23.48 \\
SSIM  & 0.8693 & 0.8535 & 0.8506 & 0.7772 & 0.8687 & 0.8521 & 0.7420 & 0.8687 & 0.8452 & 0.8667 & 0.8738 & 0.8458 & 0.7234 \\
LPIPS & 0.3293 & 0.3530 & 0.3650 & 0.3641 & 0.3486 & 0.3469 & 0.3835 & 0.3214 & 0.3436 & 0.3668 & 0.3459 & 0.3536 & 0.4121 \\
\end{tabular}
}

\resizebox{1.0\textwidth}{!}{
\begin{tabular}{cccccccccccccc}
      & suit & teddy & toast & toil & tbus & tplane & ttrain & ttruck & tv & umb & vase & wine & \textbf{overall} \\ 
\hline
PSNR  & 28.87 & 29.39 & 28.05 & 32.61 & 28.15 & 29.30 & 28.42 & 28.63 & 27.69 & 28.62 & 28.47 & 28.13 & \textbf{28.82} \\
SSIM  & 0.8609 & 0.8713 & 0.8564 & 0.9297 & 0.8549 & 0.8703 & 0.8438 & 0.8627 & 0.8709 & 0.8504 & 0.8536 & 0.8420 & \textbf{0.8564} \\
LPIPS & 0.3526 & 0.3345 & 0.3466 & 0.2841 & 0.3579 & 0.3537 & 0.3765 & 0.3546 & 0.3865 & 0.3601 & 0.3570 & 0.3820 & \textbf{0.3451} \\
\end{tabular}
}
\vspace{2mm}
\caption{Classwise rendering quality of \our CO3D dataset.}
\label{tab:supp_classwise}
\end{table}

\begin{table}[!htb]
\centering
\small
\caption{Score tables for 3D classification networks. We repeat each experiment for 3 times and report the mean ($\mu$) and standard deviation ($\sigma$). \textbf{D} denotes density and \textbf{SH} denotes spherical harmonic coefficients.}
\resizebox{.75\textwidth}{!}{
\begin{tabular}{cccccc}
\toprule
             \multicolumn{5}{c}{Acc@1 ($\mu \pm \sigma$) } \\
\midrule
Input Feature & None & \textbf{D} & \textbf{SH} & \textbf{SH + D}  \\ 
\midrule
3DResNet14    & 59.36 $\pm$ 0.30 & 72.44 $\pm$ 0.29  & 71.87 $\pm$ 0.61 & 72.92 $\pm$ 0.42 \\
3DResNet18    & 63.85 $\pm$ 0.33 & 75.18 $\pm$ 0.70 & 75.58 $\pm$ 0.37 & 75.72 $\pm$ 0.25 \\
3DResNet34    & 64.55 $\pm$ 0.84 & 76.38 $\pm$ 0.34 & 76.51 $\pm$ 0.61 & 76.50 $\pm$ 0.03 \\
3DResNet50    & 65.25 $\pm$ 0.75 & 76.42 $\pm$ 0.19 & 77.59 $\pm$ 0.17 & 77.53 $\pm$ 0.27 \\
3DResNet101   & 66.21 $\pm$ 0.93  & 77.27 $\pm$ 0.61 & 78.04 $\pm$ 0.58 & 77.19 $\pm$ 0.89 \\
\midrule
             \multicolumn{5}{c}{Acc@5 ($\mu \pm \sigma$)} \\
\midrule
Input Feature & None & \textbf{D} & \textbf{SH} & \textbf{SH + D}  \\
\midrule
3DResNet14    & 81.30 $\pm$ 0.87 & 90.04 $\pm$ 0.13  & 89.60 $\pm$ 0.36 & 90.83 $\pm$ 0.03 \\
3DResNet18    & 84.47 $\pm$ 0.54 & 91.10 $\pm$ 0.24 & 90.98 $\pm$ 0.40 & 91.54 $\pm$ 0.21 \\
3DResNet34    & 84.26 $\pm$ 0.47 & 91.79 $\pm$ 0.61 & 91.79 $\pm$ 0.49 & 91.98 $\pm$ 0.08 \\
3DResNet50    & 85.71 $\pm$ 0.64 & 92.91 $\pm$ 0.24 & 92.73 $\pm$ 0.21 & 92.93 $\pm$ 0.54 \\
3DResNet101   & 86.50 $\pm$ 0.09 & 92.68 $\pm$ 0.34 & 93.24 $\pm$ 0.49 & 93.09 $\pm$ 0.21 \\
\bottomrule
\end{tabular}
}
\label{tab:co3d_3d_cls}
\end{table}

\begin{table}[!htb]
\centering
\small
\caption{Evaluated semantic segmentation performance on \our ScanNet dataset without(left) and with(right) spherical harmonic coefficients as input feature. \textbf{SH} denotes spherical harmonic coefficients. }
\resizebox{.75\textwidth}{!}{
\begin{tabular}{cccccc}
\toprule
             \multicolumn{5}{c}{mIoU ($\mu \pm \sigma$) } \\
\midrule
Input Feature & None & \textbf{D} & \textbf{SH} & \textbf{SH + D}  \\ 
\midrule
3D-Res16UNet14A~\cite{choy20194d}   & 57.6 $\pm$ 2.0 & 60.4 $\pm$ 0.1  & 60.2 $\pm$ 0.2 & 60.3 $\pm$ 0.1\\
3D-Res16UNet18A~\cite{choy20194d}   & 59.5 $\pm$ 2.0 & 61.5 $\pm$ 0.6 & 61.8 $\pm$ 0.4 & 61.7 $\pm$ 0.5 \\
3D-Res16UNet34C~\cite{choy20194d}   & 60.2 $\pm$ 2.3 & 62.2 $\pm$ 0.9 & 62.5 $\pm$ 0.7 & 62.5 $\pm$ 0.7 \\
\midrule
             \multicolumn{5}{c}{mAcc ($\mu \pm \sigma$)} \\
\midrule
Input Feature & None & \textbf{D} & \textbf{SH} & \textbf{SH + D}  \\
\midrule
3D-Res16UNet14A~\cite{choy20194d}   & 67.9 $\pm$ 1.8 & 70.4 $\pm$ 0.0  & 69.9 $\pm$ 0.4 & 70.0 $\pm$ 0.3 \\
3D-Res16UNet18A~\cite{choy20194d}   & 69.7 $\pm$ 1.7 & 71.7 $\pm$ 0.4 & 71.7 $\pm$ 0.3 & 71.4 $\pm$ 0.4 \\
3D-Res16UNet34C~\cite{choy20194d}   & 70.1 $\pm$ 1.9 & 72.1 $\pm$ 0.5 & 72.0 $\pm$ 0.6 & 72.2 $\pm$ 0.4 \\
\bottomrule
\end{tabular}
}
\label{tab:scannet_semseg}
\end{table}

\begin{table*}[!htb]
\setlength{\tabcolsep}{2pt}
    \caption{
    Scenewise statistics of semantic segmentation networks on the \our ScanNet validation split. All experiments are performed with three different random seed values and the mean and standard deviation are reported.
    }
    \centering
    \small
    \resizebox{.99\textwidth}{!}{
    \begin{tabular}{lcccccccccccc}
    \toprule
    IoU & \textbf{D} & \textbf{SH} & bath & bed & bksf & cab & chair & cntr & curt & desk & door & floor \\
    \midrule
    14A & &  & 74.3$\pm{3.0}$ & 72.5$\pm{2.8}$ & 57.9$\pm{4.6}$ & 57.1$\pm{0.3}$ & 80.6$\pm{0.8}$ & 52.2$\pm{3.0}$ & 22.3$\pm{3.3}$ & 55.1$\pm{2.2}$ & 36.0$\pm{5.3}$ & 94.1$\pm{0.3}$ \\
    14A & \checkmark &  & 79.5$\pm{1.3}$ & 73.2$\pm{2.3}$ & 64.8$\pm{0.8}$ & 58.0$\pm{0.9}$ & 81.5$\pm{0.8}$ & 51.5$\pm{3.6}$ & 28.6$\pm{1.8}$ & 57.9$\pm{0.5}$ & 44.3$\pm{0.6}$ & 94.3$\pm{0.2}$ \\
    14A & & \checkmark & 77.0$\pm{0.9}$ & 74.5$\pm{0.7}$ & 64.6$\pm{1.0}$ & 57.6$\pm{0.7}$ & 81.5$\pm{0.2}$ & 51.8$\pm{2.9}$ & 23.7$\pm{1.7}$ & 57.9$\pm{2.5}$ & 43.3$\pm{0.3}$ & 94.5$\pm{0.0}$ \\
    14A & \checkmark & \checkmark & 76.5$\pm{1.2}$ & 75.9$\pm{0.6}$ & 64.9$\pm{1.6}$ & 57.1$\pm{1.0}$ & 82.0$\pm{0.2}$ & 50.5$\pm{2.5}$ & 22.0$\pm{2.9}$ & 57.8$\pm{1.2}$ & 43.6$\pm{0.2}$ & 94.5$\pm{0.1}$ \\
    \midrule
    18A & &  & 77.3$\pm{2.4}$ & 73.5$\pm{2.5}$ & 59.6$\pm{3.8}$ & 59.8$\pm{1.1}$ & 81.4$\pm{1.2}$ & 58.2$\pm{2.0}$ & 21.8$\pm{2.1}$ & 58.0$\pm{2.7}$ & 39.7$\pm{6.1}$ & 94.2$\pm{0.3}$ \\
    18A & \checkmark &  & 77.3$\pm{2.2}$ & 73.0$\pm{2.8}$ & 64.9$\pm{0.4}$ & 58.6$\pm{1.4}$ & 82.6$\pm{0.5}$ & 57.3$\pm{2.1}$ & 31.4$\pm{5.6}$ & 60.0$\pm{2.1}$ & 47.7$\pm{0.9}$ & 94.4$\pm{0.2}$ \\
    18A & & \checkmark & 80.6$\pm{1.0}$ & 75.3$\pm{1.6}$ & 66.1$\pm{1.5}$ & 59.3$\pm{1.4}$ & 82.0$\pm{1.0}$ & 59.6$\pm{1.1}$ & 23.9$\pm{1.8}$ & 59.9$\pm{1.2}$ & 46.9$\pm{1.0}$ & 94.6$\pm{0.0}$ \\
    18A & \checkmark & \checkmark & 80.3$\pm{3.1}$ & 75.9$\pm{0.8}$ & 66.0$\pm{1.0}$ & 59.6$\pm{1.0}$ & 82.5$\pm{0.7}$ & 55.3$\pm{3.7}$ & 26.2$\pm{2.7}$ & 60.8$\pm{0.6}$ & 46.5$\pm{1.3}$ & 94.6$\pm{0.1}$ \\
    \midrule
    34C & &  & 77.8$\pm{4.9}$ & 73.5$\pm{1.8}$ & 60.8$\pm{4.1}$ & 60.0$\pm{1.0}$ & 81.7$\pm{1.6}$ & 58.3$\pm{2.4}$ & 25.0$\pm{1.4}$ & 58.6$\pm{3.6}$ & 40.3$\pm{8.1}$ & 94.2$\pm{0.3}$ \\
    34C & \checkmark &  & 81.0$\pm{1.9}$ & 73.4$\pm{1.8}$ & 65.2$\pm{0.9}$ & 60.2$\pm{0.8}$ & 82.8$\pm{0.8}$ & 56.7$\pm{0.4}$ & 30.6$\pm{3.6}$ & 62.6$\pm{1.1}$ & 48.7$\pm{3.0}$ & 94.3$\pm{0.2}$ \\
    34C & & \checkmark & 79.9$\pm{3.1}$ & 76.3$\pm{0.3}$ & 66.2$\pm{0.3}$ & 59.6$\pm{1.3}$ & 83.7$\pm{0.4}$ & 55.9$\pm{2.7}$ & 25.8$\pm{0.4}$ & 62.2$\pm{1.5}$ & 48.6$\pm{2.9}$ & 94.6$\pm{0.1}$ \\
    34C & \checkmark & \checkmark & 79.5$\pm{2.8}$ & 75.1$\pm{0.8}$ & 66.0$\pm{0.3}$ & 60.3$\pm{1.1}$ & 83.1$\pm{0.7}$ & 57.2$\pm{0.9}$ & 24.9$\pm{1.8}$ & 61.8$\pm{1.8}$ & 49.5$\pm{1.9}$ & 94.6$\pm{0.0}$ \\
    \bottomrule
    \toprule
    IoU & \textbf{D} & \textbf{SH} & othr & pic & ref & show & sink & sofa & tab & toil & wall & wind  \\
    \midrule 
    14A & &  & 33.9$\pm{2.1}$ & 8.3$\pm{6.5}$ & 53.7$\pm{2.4}$ & 54.0$\pm{1.4}$ & 64.4$\pm{2.0}$ & 72.8$\pm{1.2}$ & 64.3$\pm{2.5}$ & 86.2$\pm{1.5}$ & 74.6$\pm{2.2}$ & 37.3$\pm{2.7}$ \\
    14A & \checkmark &  & 34.9$\pm{1.3}$ & 17.4$\pm{0.7}$ & 57.6$\pm{1.4}$ & 49.9$\pm{3.9}$ & 66.9$\pm{1.0}$ & 74.2$\pm{1.0}$ & 66.4$\pm{1.0}$ & 86.2$\pm{1.5}$ & 77.2$\pm{0.4}$ & 43.0$\pm{1.5}$ \\
    14A & & \checkmark & 35.9$\pm{0.6}$ & 17.3$\pm{1.3}$ & 58.5$\pm{1.2}$ & 52.0$\pm{2.1}$ & 66.5$\pm{0.4}$ & 74.2$\pm{0.6}$ & 66.8$\pm{0.8}$ & 88.4$\pm{1.0}$ & 77.0$\pm{0.5}$ & 39.9$\pm{0.8}$ \\
    14A & \checkmark & \checkmark & 36.5$\pm{0.7}$ & 15.3$\pm{1.5}$ & 59.2$\pm{1.5}$ & 53.4$\pm{3.0}$ & 67.3$\pm{0.7}$ & 74.7$\pm{0.7}$ & 67.1$\pm{0.8}$ & 89.0$\pm{0.5}$ & 77.1$\pm{0.4}$ & 41.8$\pm{1.0}$ \\
    \midrule
    18A & &  & 35.0$\pm{1.1}$ & 9.9$\pm{5.8}$ & 57.6$\pm{1.4}$ & 54.7$\pm{2.0}$ & 65.9$\pm{1.9}$ & 75.7$\pm{0.9}$ & 65.8$\pm{1.6}$ & 86.6$\pm{1.4}$ & 75.0$\pm{1.8}$ & 41.3$\pm{2.5}$ \\
    18A & \checkmark &  & 34.8$\pm{1.2}$ & 18.7$\pm{0.9}$ & 57.1$\pm{0.9}$ & 51.2$\pm{5.9}$ & 66.7$\pm{1.0}$ & 77.2$\pm{0.5}$ & 68.0$\pm{0.4}$ & 85.6$\pm{2.1}$ & 77.5$\pm{0.1}$ & 45.4$\pm{0.9}$ \\
    18A & & \checkmark & 36.2$\pm{0.4}$ & 16.5$\pm{1.7}$ & 58.9$\pm{1.9}$ & 56.5$\pm{1.4}$ & 66.8$\pm{1.0}$ & 74.0$\pm{1.8}$ & 68.3$\pm{0.5}$ & 89.0$\pm{1.0}$ & 77.6$\pm{0.0}$ & 43.6$\pm{1.0}$ \\
    18A & \checkmark & \checkmark & 36.4$\pm{0.3}$ & 17.2$\pm{1.6}$ & 56.7$\pm{2.9}$ & 55.5$\pm{1.6}$ & 67.1$\pm{0.9}$ & 75.5$\pm{1.7}$ & 67.5$\pm{0.4}$ & 88.7$\pm{0.8}$ & 77.6$\pm{0.2}$ & 43.1$\pm{1.4}$ \\
    \midrule
    34C & &  & 37.1$\pm{2.4}$ & 9.2$\pm{5.5}$ & 59.5$\pm{2.4}$ & 55.6$\pm{2.0}$ & 66.6$\pm{1.6}$ & 76.8$\pm{1.1}$ & 65.8$\pm{2.6}$ & 86.3$\pm{1.4}$ & 75.0$\pm{2.5}$ & 42.3$\pm{1.7}$ \\
    34C & \checkmark &  & 36.9$\pm{2.5}$ & 16.4$\pm{0.3}$ & 60.5$\pm{3.5}$ & 54.0$\pm{3.4}$ & 66.6$\pm{1.5}$ & 75.5$\pm{1.8}$ & 69.8$\pm{0.5}$ & 86.3$\pm{1.5}$ & 78.1$\pm{0.4}$ & 45.4$\pm{2.0}$ \\
    34C & & \checkmark & 37.3$\pm{2.7}$ & 17.2$\pm{1.3}$ & 61.2$\pm{2.6}$ & 57.9$\pm{1.4}$ & 67.2$\pm{1.1}$ & 77.6$\pm{0.4}$ & 69.1$\pm{0.6}$ & 88.8$\pm{0.4}$ & 78.1$\pm{0.6}$ & 43.5$\pm{0.6}$ \\
    34C & \checkmark & \checkmark & 39.7$\pm{1.3}$ & 17.0$\pm{0.5}$ & 57.8$\pm{5.6}$ & 58.2$\pm{1.7}$ & 67.8$\pm{0.7}$ & 77.2$\pm{1.6}$ & 69.2$\pm{0.1}$ & 89.3$\pm{0.7}$ & 78.2$\pm{0.2}$ & 43.6$\pm{0.5}$ \\
    \bottomrule
    \multicolumn{13}{c}{\vspace{0.3cm}}\\
    %
    \toprule
    Acc & \textbf{D} & \textbf{SH} & bath & bed & bksf & cab & chair & cntr & curt & desk & door & floor \\
    \midrule
    14A & &  & 81.6$\pm{2.3}$ & 80.8$\pm{2.0}$ & 74.1$\pm{4.7}$ & 72.3$\pm{1.2}$ & 87.7$\pm{0.7}$ & 66.4$\pm{2.3}$ & 26.7$\pm{3.3}$ & 74.0$\pm{1.6}$ & 45.5$\pm{5.6}$ & 98.1$\pm{0.2}$ \\
    14A & \checkmark &  & 86.9$\pm{1.8}$ & 80.6$\pm{1.4}$ & 79.3$\pm{1.1}$ & 73.3$\pm{0.6}$ & 88.2$\pm{0.4}$ & 64.3$\pm{2.6}$ & 33.8$\pm{3.0}$ & 75.9$\pm{0.9}$ & 55.5$\pm{1.6}$ & 98.2$\pm{0.1}$ \\
    14A & & \checkmark & 82.7$\pm{1.2}$ & 81.4$\pm{1.0}$ & 79.6$\pm{1.0}$ & 73.2$\pm{0.8}$ & 88.9$\pm{0.2}$ & 64.3$\pm{1.8}$ & 26.9$\pm{2.1}$ & 75.3$\pm{2.7}$ & 55.7$\pm{1.8}$ & 98.3$\pm{0.0}$ \\
    14A & \checkmark & \checkmark & 82.8$\pm{1.2}$ & 82.2$\pm{0.3}$ & 80.0$\pm{0.8}$ & 73.2$\pm{0.7}$ & 89.1$\pm{0.7}$ & 62.7$\pm{1.3}$ & 25.7$\pm{3.0}$ & 75.1$\pm{1.9}$ & 56.0$\pm{1.9}$ & 98.3$\pm{0.0}$ \\
    \midrule
    18A & &  & 85.8$\pm{0.4}$ & 81.7$\pm{1.0}$ & 76.0$\pm{1.9}$ & 74.3$\pm{0.3}$ & 88.5$\pm{0.4}$ & 72.1$\pm{2.7}$ & 25.6$\pm{1.4}$ & 76.3$\pm{2.6}$ & 51.3$\pm{7.4}$ & 98.2$\pm{0.0}$ \\
    18A & \checkmark &  & 83.9$\pm{2.8}$ & 81.3$\pm{1.3}$ & 78.9$\pm{0.7}$ & 74.4$\pm{0.3}$ & 89.1$\pm{0.1}$ & 70.6$\pm{3.2}$ & 37.6$\pm{8.7}$ & 77.1$\pm{2.7}$ & 62.6$\pm{0.7}$ & 98.2$\pm{0.1}$ \\
    18A & & \checkmark & 87.1$\pm{0.7}$ & 81.5$\pm{1.4}$ & 79.6$\pm{1.0}$ & 74.7$\pm{0.4}$ & 88.6$\pm{0.7}$ & 74.6$\pm{1.7}$ & 27.3$\pm{2.3}$ & 77.9$\pm{2.7}$ & 59.9$\pm{2.2}$ & 98.2$\pm{0.0}$ \\
    18A & \checkmark & \checkmark & 87.2$\pm{2.3}$ & 82.4$\pm{0.6}$ & 80.4$\pm{2.4}$ & 74.7$\pm{2.0}$ & 89.3$\pm{1.0}$ & 68.8$\pm{5.1}$ & 29.6$\pm{3.3}$ & 77.4$\pm{1.4}$ & 59.9$\pm{1.5}$ & 98.2$\pm{0.1}$ \\
    \midrule
    34C & &  & 86.1$\pm{3.7}$ & 81.4$\pm{1.6}$ & 75.6$\pm{2.3}$ & 74.6$\pm{1.7}$ & 89.3$\pm{0.6}$ & 70.3$\pm{3.2}$ & 28.7$\pm{1.7}$ & 78.1$\pm{3.1}$ & 52.9$\pm{7.6}$ & 98.1$\pm{0.2}$ \\
    34C & \checkmark &  & 87.7$\pm{1.5}$ & 81.2$\pm{0.8}$ & 78.2$\pm{2.0}$ & 74.9$\pm{1.4}$ & 89.6$\pm{0.4}$ & 71.4$\pm{1.9}$ & 36.1$\pm{5.7}$ & 81.0$\pm{1.1}$ & 60.0$\pm{3.1}$ & 98.2$\pm{0.1}$ \\
    34C & & \checkmark & 85.3$\pm{3.9}$ & 83.3$\pm{0.7}$ & 79.0$\pm{1.3}$ & 74.8$\pm{1.6}$ & 90.4$\pm{0.3}$ & 67.5$\pm{3.2}$ & 29.1$\pm{0.4}$ & 80.7$\pm{1.3}$ & 61.7$\pm{2.8}$ & 98.2$\pm{0.1}$ \\
    34C & \checkmark & \checkmark & 85.3$\pm{3.1}$ & 80.8$\pm{1.9}$ & 80.6$\pm{1.5}$ & 75.6$\pm{0.9}$ & 89.8$\pm{0.5}$ & 71.4$\pm{1.9}$ & 28.0$\pm{2.0}$ & 80.9$\pm{1.8}$ & 61.4$\pm{1.4}$ & 98.2$\pm{0.1}$ \\
    \bottomrule
    \toprule
    Acc & \textbf{D} & \textbf{SH} & othr & pic & ref & show & sink & sofa & tab & toil & wall & wind \\
    \midrule
    14A & &  & 41.1$\pm{1.6}$ & 10.7$\pm{8.7}$ & 65.6$\pm{1.5}$ & 64.4$\pm{0.8}$ & 76.6$\pm{0.8}$ & 86.9$\pm{0.6}$ & 75.0$\pm{2.6}$ & 92.3$\pm{0.1}$ & 91.7$\pm{1.0}$ & 45.9$\pm{4.5}$ \\
    14A & \checkmark &  & 41.9$\pm{1.0}$ & 22.2$\pm{0.7}$ & 67.9$\pm{0.4}$ & 59.2$\pm{4.3}$ & 75.5$\pm{1.1}$ & 87.5$\pm{0.3}$ & 78.8$\pm{1.4}$ & 92.9$\pm{0.4}$ & 92.3$\pm{0.5}$ & 54.2$\pm{1.5}$ \\
    14A & & \checkmark & 43.0$\pm{0.6}$ & 21.5$\pm{1.9}$ & 68.3$\pm{0.9}$ & 61.6$\pm{3.8}$ & 76.1$\pm{0.4}$ & 87.8$\pm{0.5}$ & 77.9$\pm{0.8}$ & 92.6$\pm{0.4}$ & 92.5$\pm{0.4}$ & 50.0$\pm{2.3}$ \\
    14A & \checkmark & \checkmark & 43.3$\pm{1.0}$ & 19.3$\pm{2.6}$ & 68.2$\pm{1.6}$ & 63.3$\pm{1.9}$ & 75.9$\pm{0.4}$ & 88.5$\pm{1.3}$ & 78.5$\pm{0.5}$ & 93.1$\pm{0.5}$ & 92.3$\pm{0.5}$ & 52.6$\pm{1.4}$ \\
    \midrule
    18A & &  & 42.5$\pm{1.7}$ & 11.7$\pm{7.2}$ & 69.1$\pm{2.4}$ & 64.5$\pm{2.4}$ & 75.7$\pm{1.5}$ & 87.7$\pm{2.0}$ & 77.3$\pm{2.0}$ & 92.0$\pm{0.3}$ & 90.9$\pm{1.2}$ & 52.5$\pm{3.8}$ \\
    18A & \checkmark &  & 42.4$\pm{1.4}$ & 24.3$\pm{1.8}$ & 67.3$\pm{3.0}$ & 61.1$\pm{6.7}$ & 76.4$\pm{0.8}$ & 88.9$\pm{1.1}$ & 78.8$\pm{2.4}$ & 92.5$\pm{0.5}$ & 91.2$\pm{0.8}$ & 58.1$\pm{3.0}$ \\
    18A & & \checkmark & 43.5$\pm{1.2}$ & 20.3$\pm{1.8}$ & 69.5$\pm{0.8}$ & 65.4$\pm{2.5}$ & 76.0$\pm{1.2}$ & 88.2$\pm{0.6}$ & 80.6$\pm{1.8}$ & 92.5$\pm{0.5}$ & 92.1$\pm{0.3}$ & 55.7$\pm{2.6}$ \\
    18A & \checkmark & \checkmark & 43.5$\pm{0.9}$ & 21.0$\pm{2.0}$ & 69.6$\pm{2.2}$ & 63.9$\pm{3.4}$ & 76.4$\pm{0.7}$ & 87.9$\pm{0.2}$ & 79.0$\pm{0.4}$ & 92.4$\pm{0.4}$ & 92.3$\pm{0.2}$ & 54.8$\pm{2.9}$ \\
    \midrule
    34C & &  & 44.6$\pm{1.9}$ & 11.4$\pm{7.4}$ & 72.2$\pm{4.3}$ & 63.1$\pm{2.5}$ & 76.8$\pm{0.8}$ & 86.9$\pm{2.7}$ & 75.9$\pm{2.0}$ & 91.8$\pm{0.5}$ & 90.8$\pm{1.5}$ & 53.8$\pm{2.9}$ \\
    34C & \checkmark &  & 43.4$\pm{2.8}$ & 22.5$\pm{1.8}$ & 70.9$\pm{3.0}$ & 61.2$\pm{3.7}$ & 77.1$\pm{1.2}$ & 86.7$\pm{3.0}$ & 80.2$\pm{1.1}$ & 92.8$\pm{0.5}$ & 92.8$\pm{0.2}$ & 55.6$\pm{2.8}$ \\
    34C & & \checkmark & 45.0$\pm{1.9}$ & 21.1$\pm{1.9}$ & 71.3$\pm{0.8}$ & 66.6$\pm{4.6}$ & 76.9$\pm{0.6}$ & 89.7$\pm{0.8}$ & 79.3$\pm{0.7}$ & 92.6$\pm{0.3}$ & 92.6$\pm{0.3}$ & 54.7$\pm{1.4}$ \\
    34C & \checkmark & \checkmark & 46.6$\pm{1.4}$ & 21.4$\pm{0.2}$ & 72.0$\pm{1.9}$ & 66.5$\pm{0.9}$ & 76.7$\pm{1.0}$ & 88.6$\pm{1.3}$ & 78.9$\pm{0.3}$ & 92.7$\pm{0.3}$ & 92.7$\pm{0.3}$ & 56.0$\pm{1.8}$ \\
    \bottomrule
    \end{tabular}
    }
    \label{tbl:scannet_scenewise}
\end{table*}

\subsection{Qualitative Results}

\begin{figure*}[!htb]
\centering
\includegraphics[width=\textwidth]{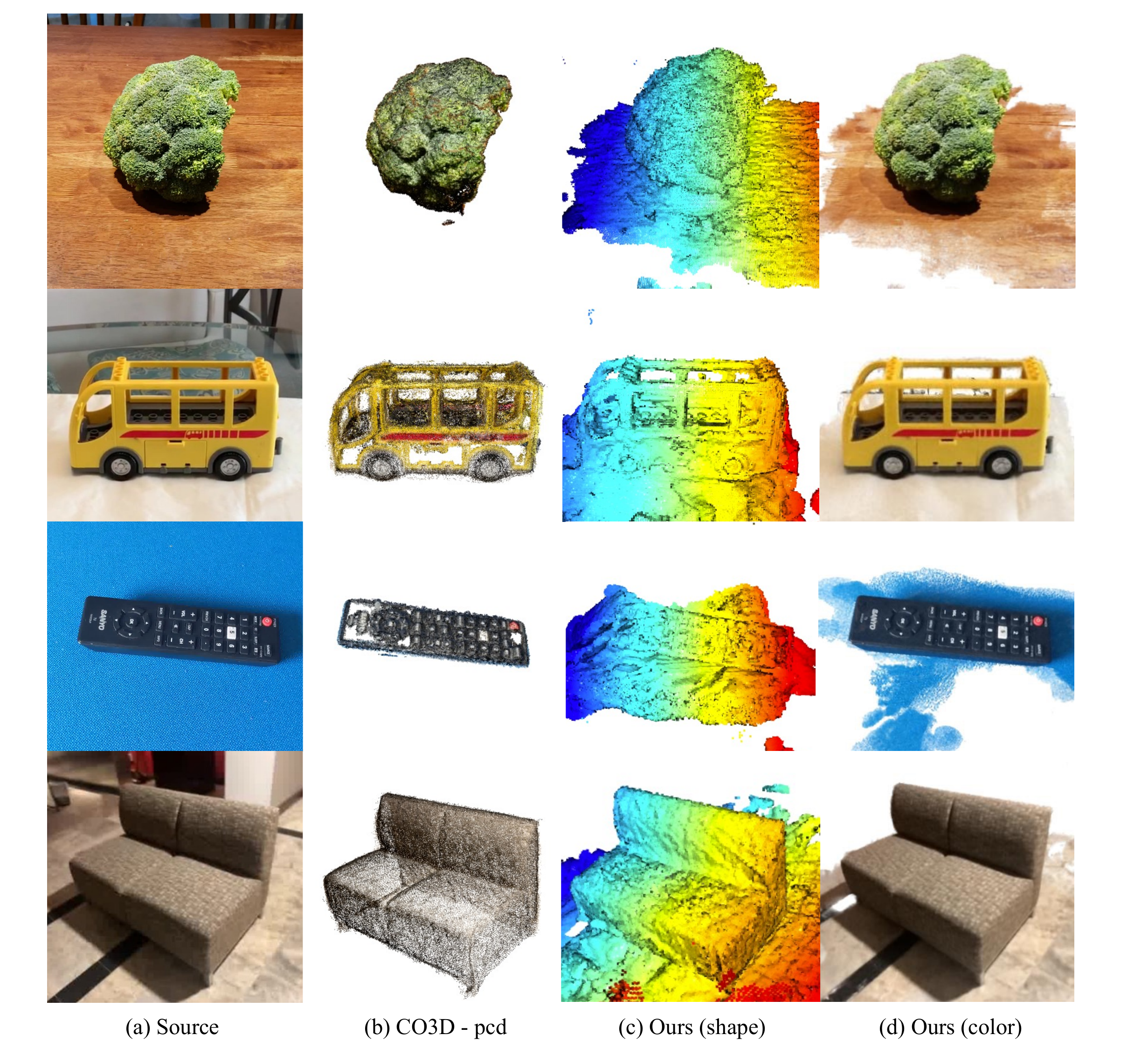}
\caption{Visualization of a few example data of original CO3D and \our CO3D.}
\label{fig:supp_data_visualize}
\end{figure*}

\begin{figure*}[!htb]
\centering
\includegraphics[width=\textwidth]{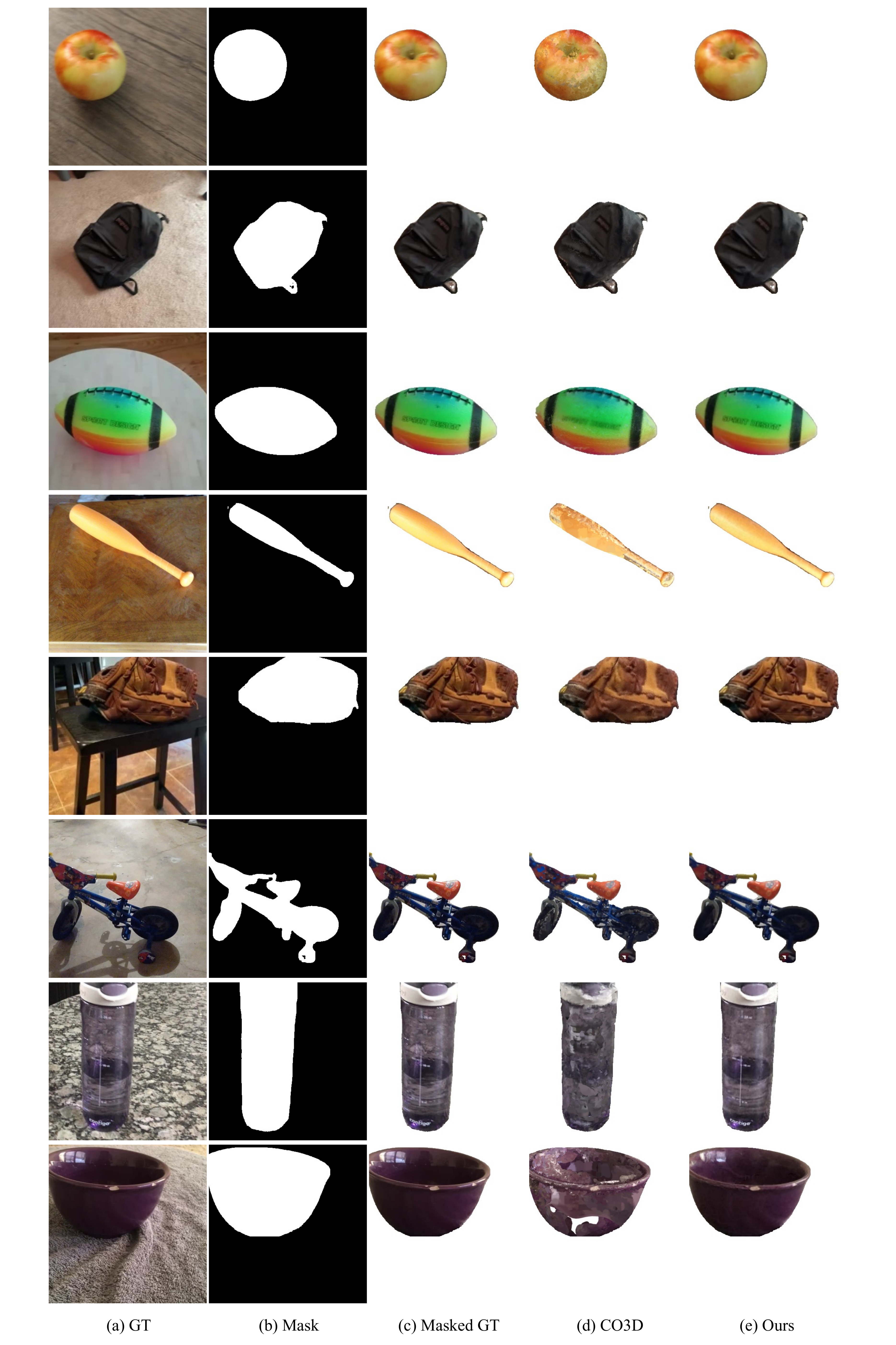}
\caption{Comparing visual reconstruction quality of original CO3D and \our CO3D.}
\label{fig:mesh_recon_1}
\end{figure*}

\begin{figure*}[!htb]
\centering
\includegraphics[width=\textwidth]{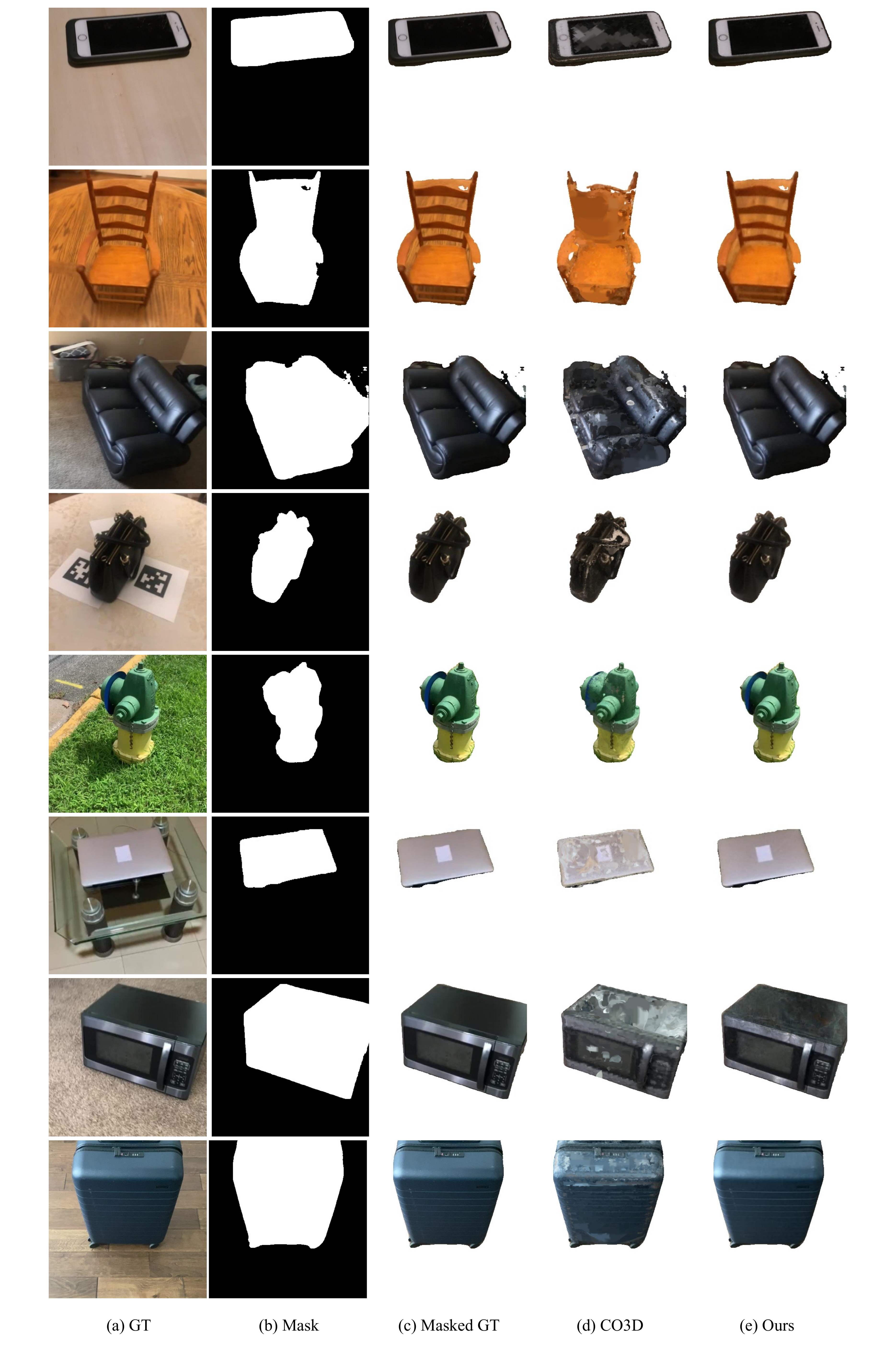}
\caption{Comparing visual reconstruction quality of original Co3D and \our Co3D.}
\label{fig:mesh_recon_2}
\end{figure*}

\begin{figure*}[!htb]
\centering
\includegraphics[width=\textwidth]{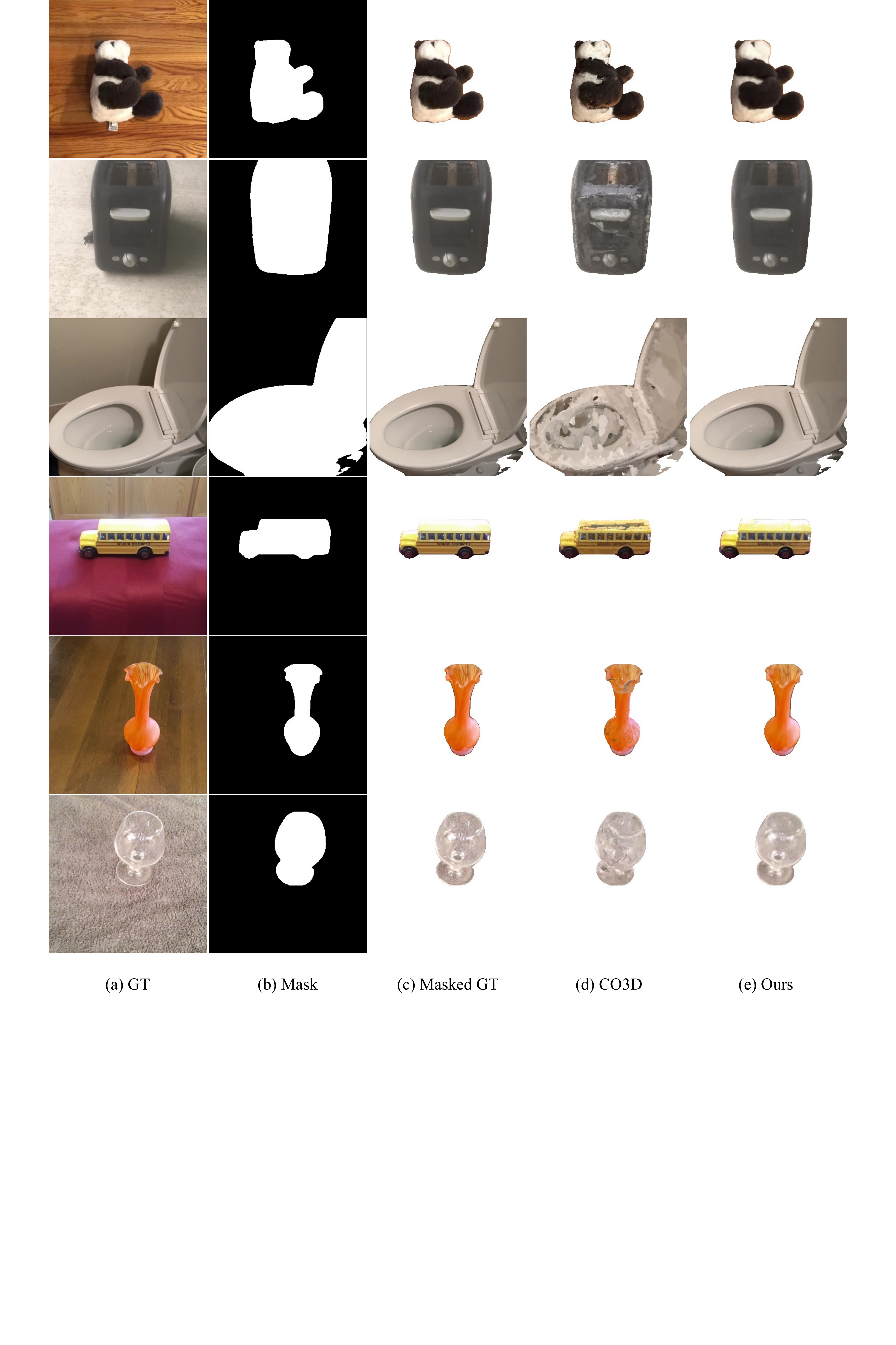}
\caption{Comparing visual reconstruction quality of original Co3D and \our Co3D.}
\label{fig:mesh_recon_3}
\end{figure*}

\begin{figure}[!htb]
\centering
\includegraphics[width=\textwidth]{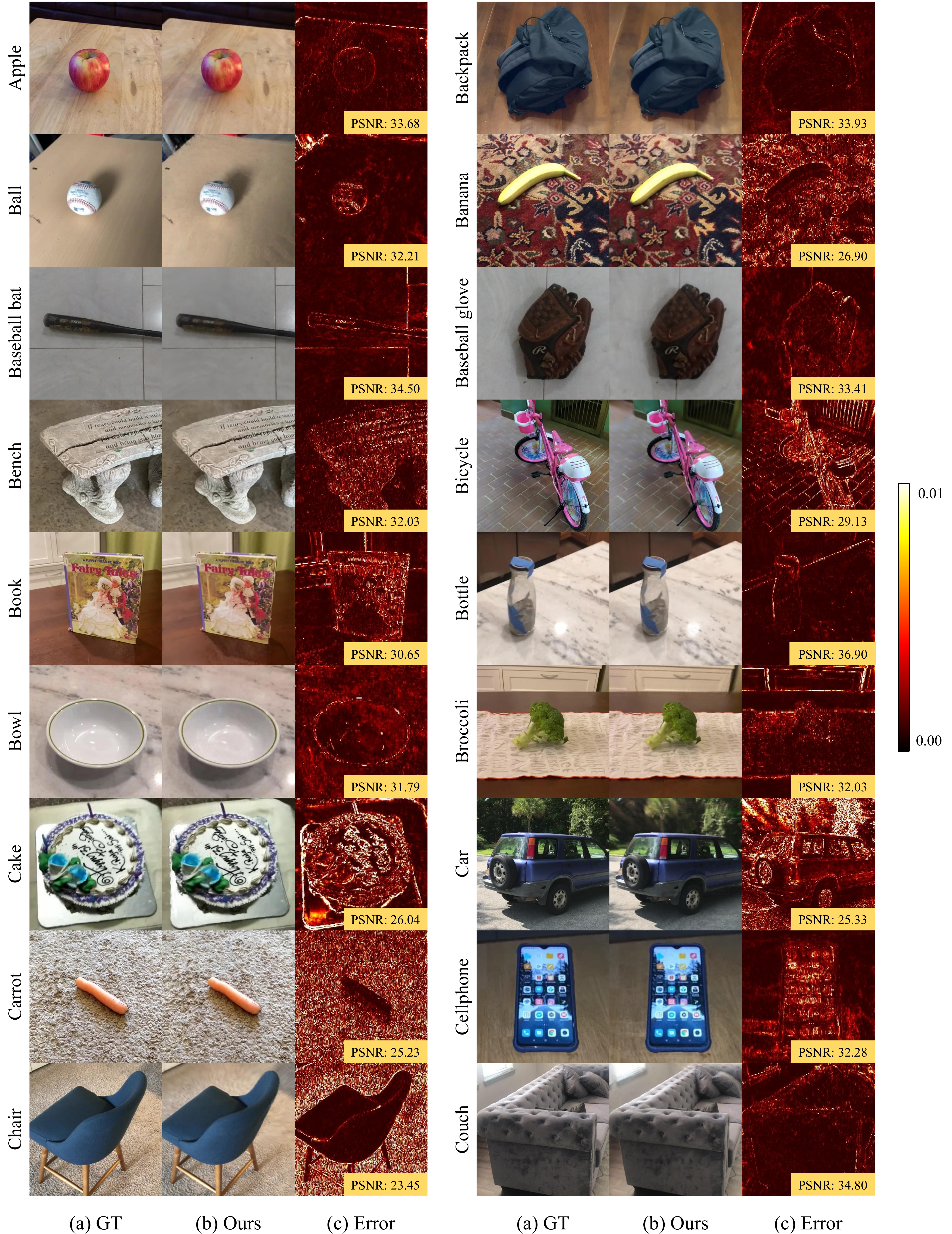}
\caption{Rendered class-wise novel views of \our CO3D. The number in error maps denote the estimated PSNR.}
\vspace{-4.0mm}
\label{fig:supp_novel_view_1}
\end{figure}

\begin{figure}[!tb]
\centering
\includegraphics[width=\textwidth]{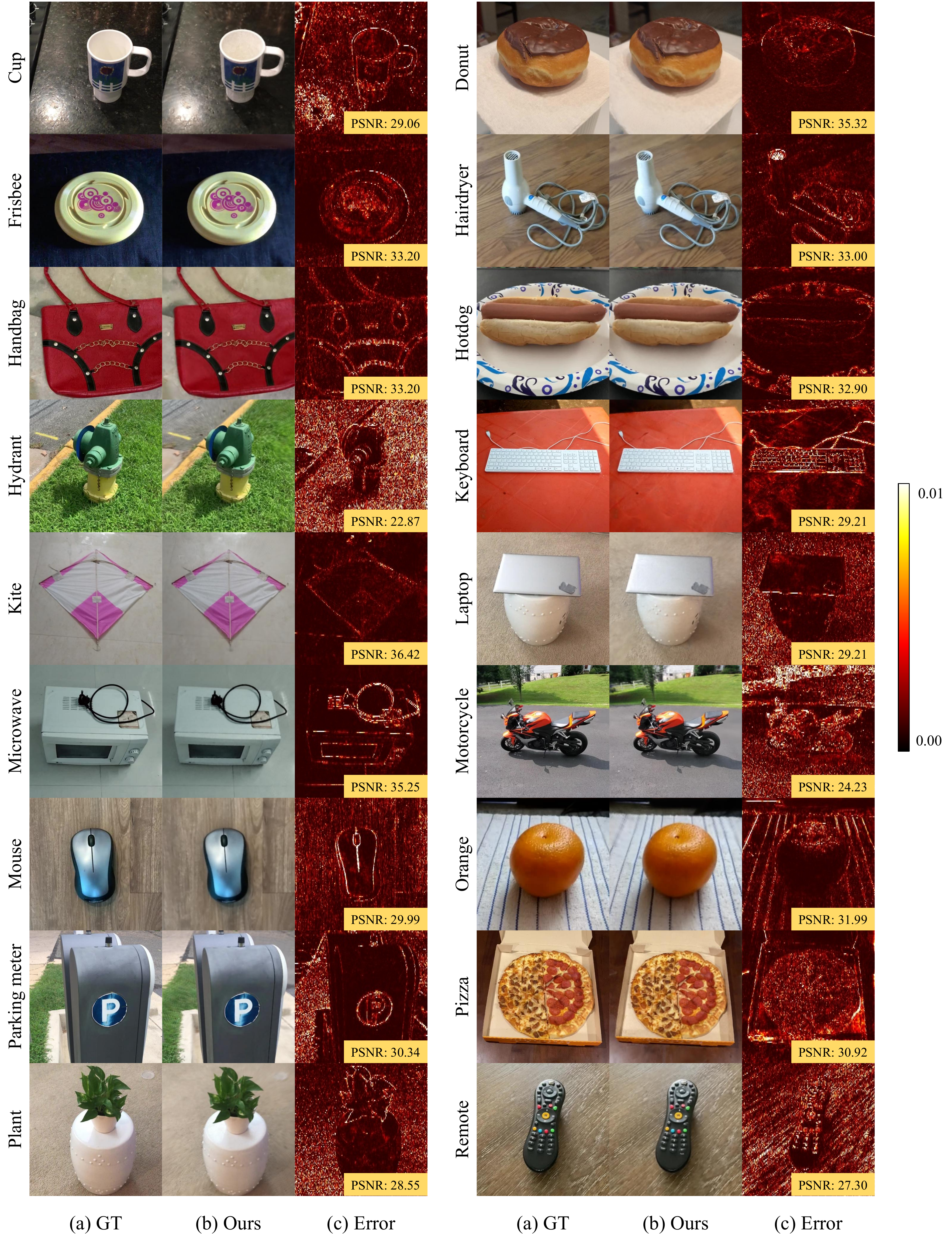}
\caption{Rendered class-wise novel views of \our CO3D. The number in error maps denote the estimated PSNR.}
\vspace{-4.0mm}
\label{fig:supp_novel_view_2}
\end{figure}

\begin{figure}[!tb]
\centering
\includegraphics[width=\textwidth]{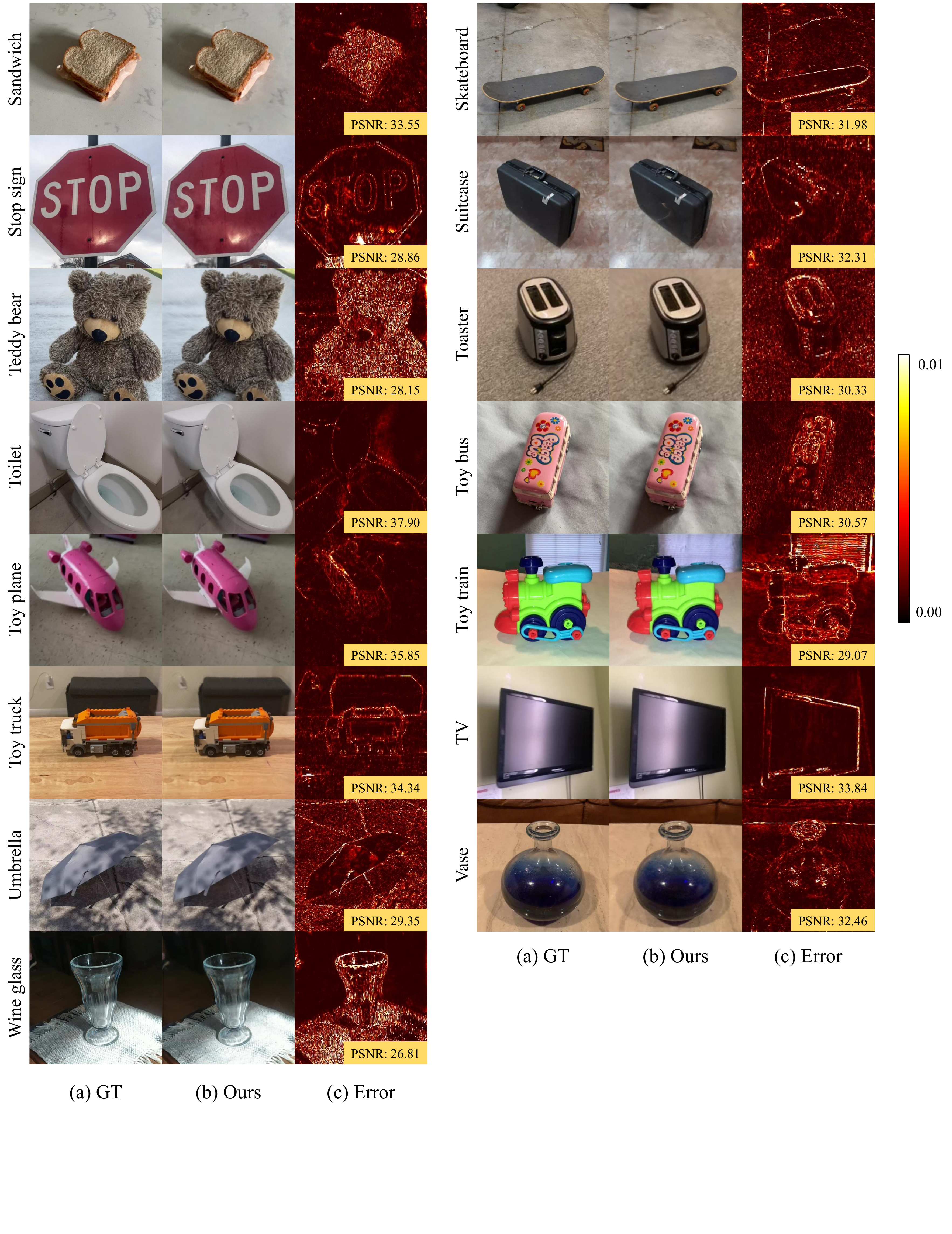}
\caption{Rendered class-wise novel views of \our CO3D. The number in error maps denote the estimated PSNR.}
\vspace{-4.0mm}
\label{fig:supp_novel_view_3}
\end{figure}

\begin{figure*}[!htb]
\centering
\includegraphics[width=0.995\textwidth]{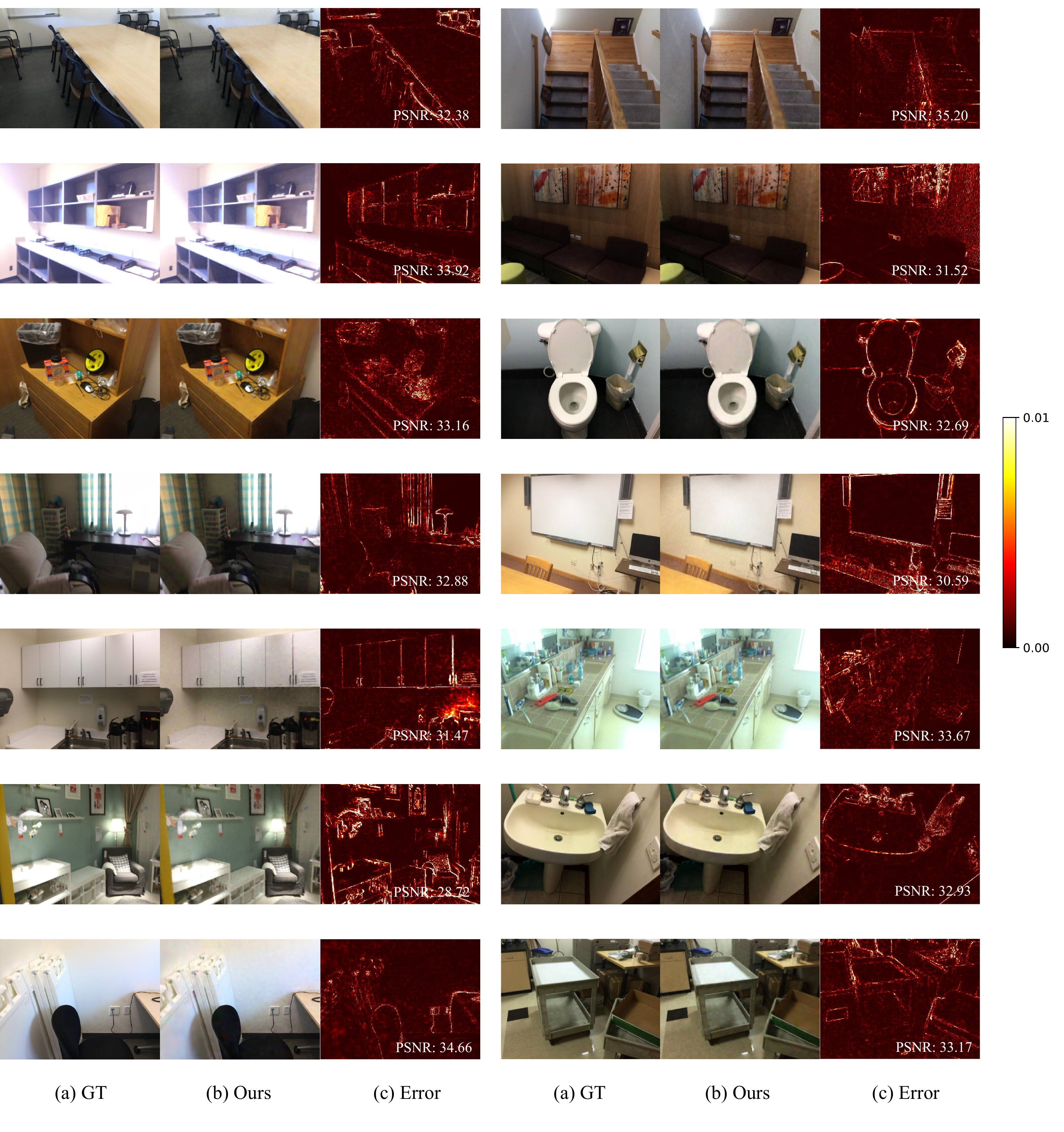}
\caption{Rendered class-wise novel views of \our ScanNet. The number in error maps denote the estimated PSNR}
\label{fig:scannet_render_qual_supp}
\end{figure*}

\begin{figure*}[!htb]
\centering
\includegraphics[width=0.995\textwidth]{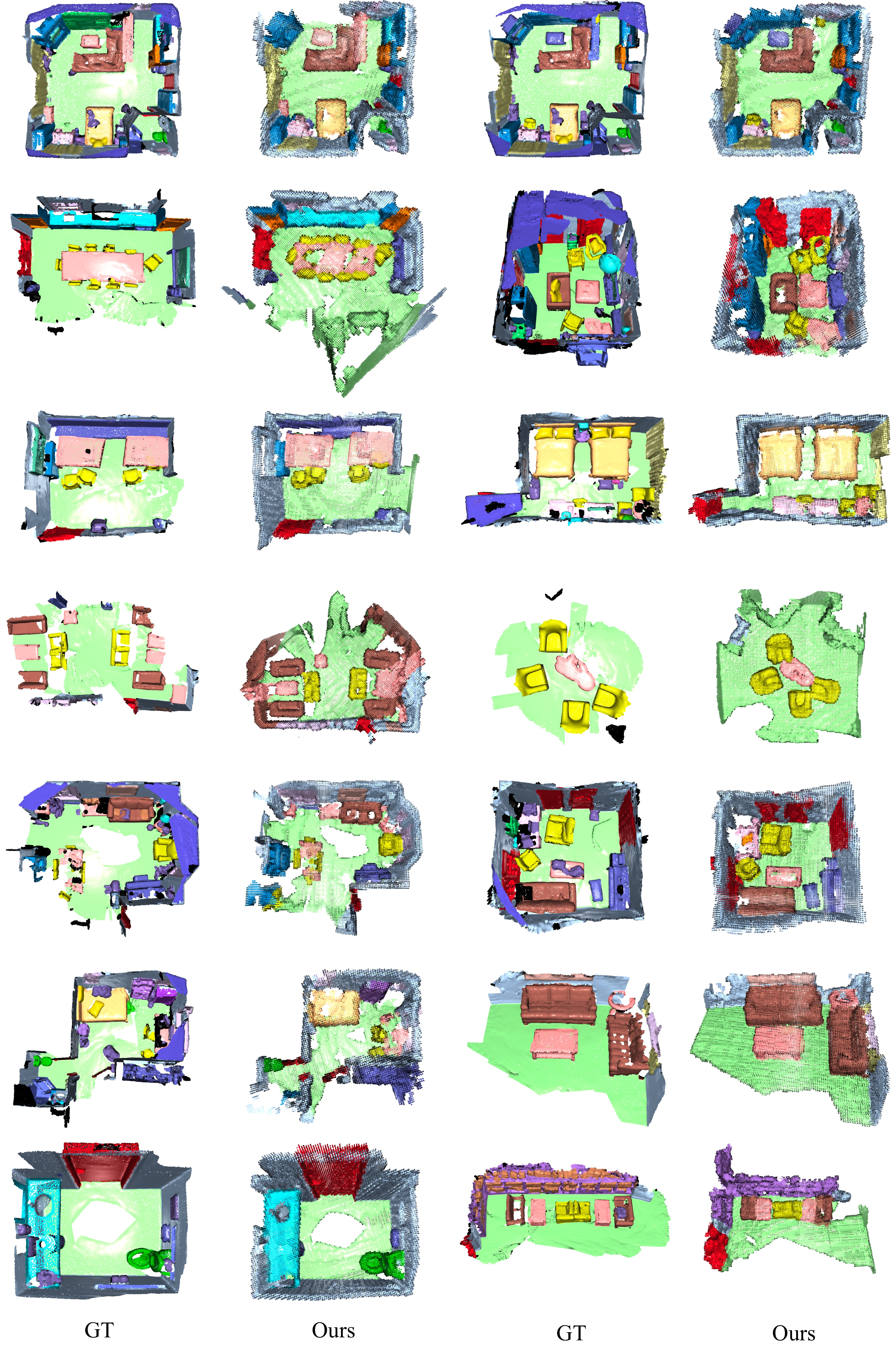}
\caption{Qualitative results of semantic segmentation on \our ScanNet dataset. (1st, 3rd columns) Ground truth point cloud with ground truth semantic labels, (2nd, 4th columns) Reconstructed sparse voxels with predicted semantic labels }
\label{fig:scannet_qual_supp}
\end{figure*}

We provide additional qualitative results introduced in the main paper. 
Figure~\ref{fig:supp_data_visualize} illustrates several examples of \our CO3D with accurate geometric information and photorealistic rendering quality.
Figure~\ref{fig:mesh_recon_1},~\ref{fig:mesh_recon_2} and~\ref{fig:mesh_recon_3} compare visual reconstruction ability of CO3D and \our CO3D. Figure~\ref{fig:supp_novel_view_1},~\ref{fig:supp_novel_view_2}, and~\ref{fig:supp_novel_view_3} visualize rendered novel views and their corresponding error maps.
In Figure~\ref{fig:scannet_render_qual_supp}, we provide the qualitative results of rendered novel views and their corresponding error maps on \our ScanNet dataset. 
Figure~\ref{fig:scannet_qual_supp} visualizes reconstructed sparse voxels of \our ScanNet with predicted semantic labels.
We have attached an additional video to show our photorealistic rendering.


\end{document}